\documentclass[11pt, a4paper, twocolumn, copyright, gdm]{google}

\usepackage[authoryear, sort&compress, round]{natbib}
\usepackage{epigraph}
\usepackage{multirow}
\usepackage[most]{tcolorbox}
\usepackage{newtxmath}
\usepackage{xspace}
\usepackage{cleveref}

\newcommand{\highlight}[1]{\cellcolor{blue!10}{#1}}

\newtcblisting{promptblock}[1]{
    colback=gray!2,
    colframe=gray!65!black,
    title=#1,
    breakable,
    listing only,
    listing options={
        breaklines=true,
        basicstyle=\ttfamily\small,
        breakatwhitespace=true
    }
}

\newtcblisting{agentbox}[0]{
    colback=gray!2,
    colframe=blue!75!black,
    title=Agent,
    breakable,
    listing only,
    listing options={
        breaklines=true,
        basicstyle=\ttfamily\small,
        columns=fullflexible,
        breakatwhitespace=false
    }
}

\newtcblisting{envbox}[0]{
    colback=gray!2,
    colframe=red!75!black,
    title=Environment,
    breakable,
    listing only,
    listing options={
        breaklines=true,
        basicstyle=\ttfamily\small,
        breakatwhitespace=true
    }
}
\newcommand{\name}{\textsc{MaD} Physics\xspace}

\uselogo{} 

\title{\name: Evaluating information seeking under constraints in physical environments}

\correspondingauthor{mokshjn00@gmail.com, nenadt@google.com}

\author[1,2,*]{Moksh Jain}
\author[1]{Mehdi Bennani}
\author[1]{Johannes Bausch}
\author[1]{Yuri Chervonyi}
\author[1]{Bogdan Georgiev}
\author[1]{Simon Osindero}
\author[1]{Nenad Toma\v{s}ev}

\affil[1]{\thepa{}{}}
\affil[2]{Mila -- Quebec AI Institute, Universit\'e de Montr\'eal}
\affil[*]{Work done during an internship at Google DeepMind}

\begin{abstract}
Scientific discovery is fundamentally a resource-constrained process that requires navigating complex trade-offs between the quality and quantity of measurements due to physical and cost constraints. Measurements drive the scientific process by revealing novel phenomena to improve our understanding. 
Existing benchmarks for evaluating agents for scientific discovery focus on either static knowledge-based reasoning or unconstrained experimental design tasks, and do not capture the ability to make measurements and plan under constraints. To bridge this gap, we propose Measuring and Discovering Physics (\name), a benchmark to evaluate the ability of agents to make informative measurements and conclusions subject to constraints on the quality and quantity of measurements. The benchmark consists of three environments, each based on a distinct physical law. To mitigate contamination from existing knowledge, \name includes altered physical laws. In each trial, the agent makes measurements of the system until it exhausts an allotted budget and then the agent has to infer the underlying physical law to make predictions about the state of the system in the future. \name evaluates two fundamental capabilities of scientific agents: inferring models from data and planning under constraints. We also demonstrate how \name can be used to evaluate other capabilities such as multimodality and in-context learning. We benchmark agents on \name using four Gemini models (2.5 Flash Lite, 2.5 Flash, 2.5 Pro, and 3 Flash), identifying shortcomings in their structured exploration and data collection capabilities and highlighting directions to improve their scientific reasoning.

\end{abstract}

\begin{document}

\maketitle

\epigraph{\textit{``Measure what can be measured, and make measurable what cannot be measured.''}}{-- Galileo}

\section{Introduction}

Scientific discovery is fundamentally a process of extracting information from the physical world \emph{under constraints}. Whether determining the position of a celestial body or measuring the properties of a quantum state, scientists constantly navigate trade-offs between the quality of measurements and their cost in time, energy, or money. While the importance of \emph{interventional} experiments, where a scientist actively manipulates a system, is well-recognized~\citep{ryan2016review,majumder2024data,m2024augmenting}, the complementary ability to make strategic, \emph{passive} observations is equally critical, particularly when interventions are impossible or prohibitively expensive.

The quest to use AI systems to accelerate the scientific process has been a central theme throughout the development of artificial intelligence~\citep{langley1981data,lloyd2014automatic,majumder2024data}. However, the most impactful contributions have been limited to specialized machine learning models targeting specific parts of the scientific process~\citep{jumper2021highly}. Recent advances in systems based on large language models (LLMs)~\citep{google2025gemini,openai2025gpt5} have demonstrated remarkable reasoning abilities on a variety of challenging problems, and have revived interest in autonomous scientific agents~\citep{lu2024ai, zheng2025automation, yamada2025ai}. LLM-based systems have already shown promising results in accelerating scientific progress~\citep{romera2024mathematical,gottweis2025towards,novikov2025alphaevolve,bubeck2025early}. However, principled evaluation of these abilities of agents remains a challenge. Existing benchmarks for scientific discovery often focus on static question-answering based on existing knowledge~\citep{rein2024gpqa,wang2025frontierscience,phan2026hle}, or on environments where agents can perform interventional experiments without any constraints~\citep{nagele2025agentic,cerrato2026sciencegym}. Additionally, these benchmarks typically rely on standard, well-known physical laws. This allows agents to succeed simply by retrieving memorized knowledge rather than through a process of empirical discovery. Furthermore, they often neglect the practical reality of science: that data acquisition is expensive, and resources must be allocated to maximize information gain.

To bridge this gap, we introduce Measuring and Discovering Physics (\name), a benchmark designed to evaluate the ability of agents to make informative measurements in physical systems under strict budget constraints. \name simulates the challenge of an experimentalist who must characterize a novel system. Each task places the agent in a dynamical environment governed by physical laws that may be ``altered'' or distinct from standard physics, such as classical mechanics with anisotropic inertia, quantum systems with modified probability rules, or fluids with non-standard forcing terms. This ensures that the agent cannot rely solely on existing knowledge but must engage in active discovery.

In \name, the agent operates in a loop of \textit{measurement} and \textit{prediction}. It is allocated a fixed budget and must choose \textit{what} to measure (e.g., which particle, which location), \textit{when} to measure it, and with \textit{what fidelity} (trading off cost for precision). Once the budget is exhausted, the agent must leverage its collected observations to infer the underlying dynamics and predict the future state of the system. This setup evaluates two fundamental capabilities: the strategic planning of measurements to resolve uncertainty, and the ability to infer complex models from noisy, sparse data.

\begin{figure}[ht]
    \centering
    \includegraphics[width=0.95\linewidth]{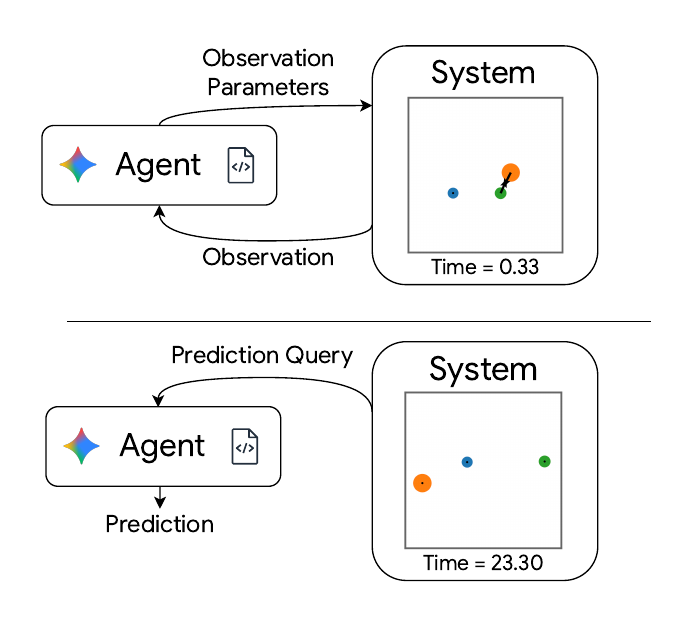}
    \caption{\name consists of two phases: Measurement (\textbf{Top}), and Prediction (\textbf{Bottom}). In the measurement phase, the agent interacts with the environment to make observations, and each observation incurs a cost. In the prediction phase, the agent has to use the observations collected during the measurement phase to predict the state of the system at the queried time.}
    \label{fig:illustrations}
\end{figure}

Our contributions are summarized below:
\begin{itemize}
    \item We propose a novel benchmarking methodology for evaluating information-seeking agents in physical domains, emphasizing the cost-fidelity trade-off in experimental design.
    \item We present a suite of three parametric environments—covering Classical, Quantum, and Fluid mechanics—that include ``altered'' physical laws to rigorously test an agent's ability to discover simplified but novel dynamics.
    \item We benchmark Gemini 2.5 Flash Lite, Flash and Pro as well as Gemini 3 Flash, on \name. We analyze their performance across different degrees of alteration and demonstrate how strategic prompting can improve their ability to navigate scientific uncertainty.
\end{itemize}

\section{Related Work}

\paragraph{Active sensing and adaptive experimental design.} The task setup used in \name is closely related to the problem of active sensing~\citep{yang2016theoretical,veiga2023reactive} and adaptive experimental design~\citep{Lindley1956OnAM,ryan2016review,rainforth2024modern}. Active sensing studies the problem of sequentially choosing informative measurements with the goal of inferring an underlying parameter with the assumption that the measurements do not affect the data generating process. On the other hand, experimental design studies the problem of designing experiments with the goal of inferring parameters for a model of the data generating process, where the experiments control parameters of the data generating process. While recent work has studied some problems at scale~\citep{hamilton2025active}, problems used within active sensing and experimental design are generally low-dimensional, and operate under the assumption that the structural form of the model is known. Recent work has also started extending active sensing to richer physical settings: LASER~\citep{deng2026laser} formulates active sensing for continuum field reconstruction as a POMDP and uses a latent world model to provide intrinsic reward signals for sensor motion. Additionally, recent work has also studied various extensions with multi-fidelity observations~\citep{garnett_bayesoptbook_2023} and look-ahead planning~\citep{jiang2020binoculars} but the tasks remain limited to simple low-dimensional observations and do not consider complex design spaces. \name does not provide the structural form of the model, requires processing high-dimensional observations and planning with cost-fidelity tradeoff for measurements.   

\paragraph{Symbolic regression.} Another problem related to the setup we study is that of symbolic regression, where the goal is to infer the symbolic form of a function given access to a dataset of inputs and corresponding outputs~\citep{koza1994genetic,schmidt2009distilling,brunton2016discovering, udrescu2020ai}. Popular approaches for symbolic regression use evolutionary algorithms to search in the space of symbolic expressions~\citep{cranmer2023interpretable} or policies trained with reinforcement learning to produce the symbolic expression that fits the data best~\citep{petersen2021deep,la2021contemporary}. Recent efforts have also explored using the domain knowledge in LLMs to improve symbolic regression~\citep{holt2024data,shojaee2025llmsrbench}. The standard symbolic regression setup assumes access to a dataset of input-output pairs from which the function is to be deciphered. Recent work has extended the setup to include acquisition of the data as part of the setup. NewtonBench~\citep{zheng2025newtonbench} and PhysMaster~\citep{miao2025physmaster} evaluate the ability of agents to perform experiments to acquire data and then infer the underlying physical laws. In contrast to measuring the exact symbolic accuracy, which is standard in symbolic regression, \name focuses on measuring the prediction accuracy. In the context where the agent is provided a limited budget to acquire data, there can be multiple symbolic expressions that fit the collected data equally well and prediction accuracy becomes more critical.

\paragraph{Evaluating agents for scientific discovery.} Existing benchmarks for scientific discovery typically focus on abilities such as hypothesis generation, experiment design, literature review or static knowledge-based reasoning~\citep{melnik2023benchmarks}. Benchmarks such as GPQA~\citep{rein2024gpqa}, FrontierScience~\citep{wang2025frontierscience}, PHYBench~\citep{qiu2025phybench}, Humanity's Last Exam (HLE)~\citep{phan2026hle} and CritPT~\citep{zhu2025probing} consist of question-answering tasks which require the knowledge about the topic as well as reasoning to arrive at the correct answer. PhysicsMind~\citep{mak2026physicsmind} is a visual question-answering dataset to evaluate the physical understanding of VLMs. SymPyBench~\citep{imani2025sympybench} consists of parameterized physics questions paired with different types of responses.  Question-answer based benchmarks evaluate knowledge and reasoning but do not measure the ability of agents to dynamically collect data, make inferences from it and plan under constraints. 

Among interactive benchmarks, ScienceWorld~\citep{wang2022scienceworld}, DiscoveryWorld~\citep{jansen2024discoveryworld}, and Alchemy~\citep{wang2021alchemy} frame scientific discovery tasks as a game with a goal where successfully achieving the goal requires understanding the environment dynamics. AutumnBench~\citep{autumnbench} consists of environments on a grid where the agent acts in the absence of a goal to understand the environment dynamics and is then tested on various tasks that require understanding the environment dynamics. DiscoveryBench~\citep{majumder2025discoverybench} and AstaBench~\citep{bragg2025astabenchrigorousbenchmarkingai} evaluate agents holistically throughout the entire scientific discovery pipeline, including literature review, code implementation and execution, hypothesis generation and experimental design. PhysGym~\citep{chen2025physgym} transforms questions from PHYBench~\citep{qiu2025phybench} into interactive environments where the agent can run the system with various controllable interventions and the goal is to infer the exact symbolic expression governing the system. \citet{yang2026think} proposed an agent augmented with physics-based tools for this symbolic discovery setting. In contrast, \name focuses on the setting where the agent cannot alter the system and is only allowed to make measurements, and each observation is associated with a cost based on the quality of the measurement chosen. 

GravityBench~\citep{koblischke2025gravitybenchv}, closely related to \name, consists of 2-body gravitational simulations where the agent is provided a budget for the number of observations allowed, after which the agent is required to answer questions with numerical or binary True-False answers. Complementary to GravityBench, \name consists of 3 different physical domains and involves reasoning about the cost-quality tradeoff for observations.

\citet{sawyer2024can} evaluate the ability of foundation model based agents to actively gather information in Alchemy~\citep{wang2021alchemy} and observe strong exploratory capabilities in simple environments but note that agents struggle to learn over multiple trials. \citet{nagele2025agentic} evaluate the ability of agents to solve problems by exploring computational models of physical systems, allowing the agent to run the system with different configurations and the ability to observe the entire simulation outcome and is required to output a function which produces the correct answer.

\paragraph{Agents for scientific discovery.} Alongside benchmark development, recent work has proposed domain-specific agentic systems for scientific discovery. PiFlow~\citep{pu2025piflow} studies principle-aware multi-agent collaboration for structured scientific exploration, PhysMaster~\citep{miao2025physmaster} targets autonomous theoretical and computational physics research, and AI Agents for Variational Quantum Circuit Design~\citep{knipfer2026vqcagents} applies agentic search to quantum circuit architecture design. These systems complement benchmark efforts by illustrating concrete discovery workflows in specific domains, whereas \name is designed as an evaluation framework for information gathering and inference under observation constraints in physical environments.

\begin{figure*}
    \centering

    \includegraphics[width=\linewidth]{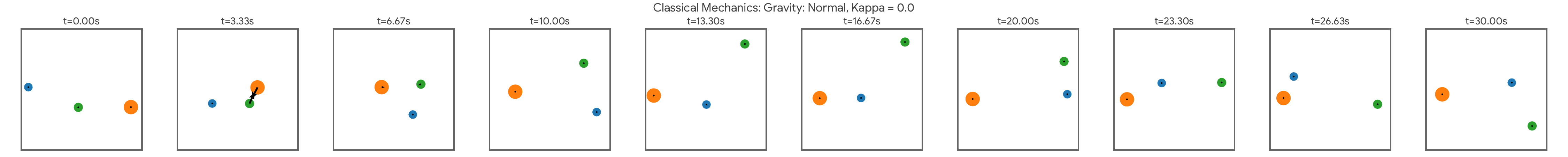}
    \includegraphics[width=\linewidth]{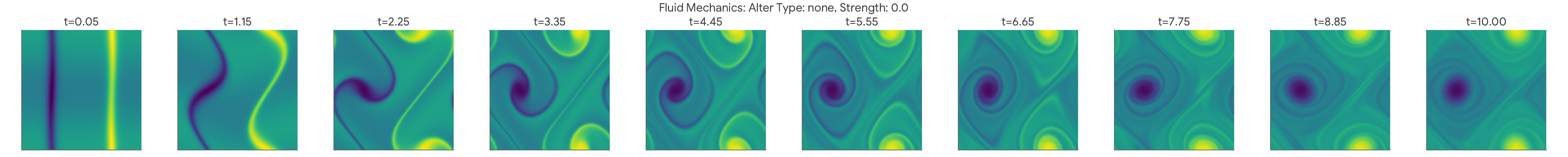}
    \includegraphics[width=\linewidth]{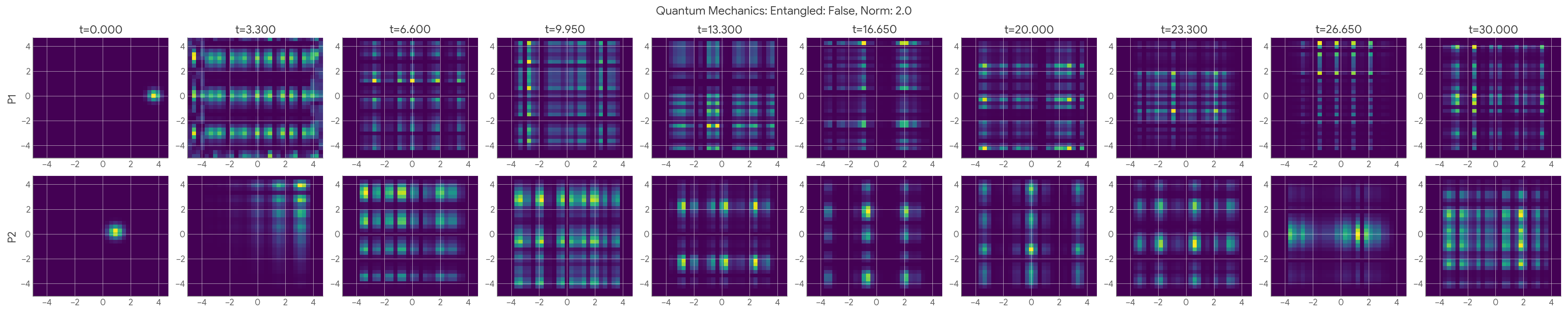}
    \caption{Visualizations of the systems used in the Classical (\textbf{Top}), Fluid (\textbf{Middle}), and Quantum (\textbf{Bottom}) mechanics environments.}
    \label{fig:env_examples}
\end{figure*}

\section{Environment Design}
\label{sec:env_design}

\subsection{Setup}
We consider a time-dependent dynamical system defined by a generalized state variable $s(t)$. This state evolves according to a governing equation:
\begin{equation}
    \partial_t s(t) = F(s(t), t), \quad s(0) = s_0
\end{equation}
where $F$ represents the physical laws driving the evolution and $s_0$ is the initial configuration. The specific nature of $s(t)$ depends on the physical domain.

The agent interacts with the system through discrete measurements and predictions. The agent does not receive a continuous stream of data but must strategically choose distinct observations to maximize information gain within a fixed budget. The evaluation protocol for the agent consists of two distinct phases.

\paragraph{Measurement.}
In the measurement phase, the agent operates under a total budget $B$ and interacts with the environment sequentially to collect a set of observations. Each observation choice is defined by the tuple $(t_k, o_k, \sigma_k)$, where:
\begin{itemize}
    \item $t_k$ is the selected time of measurement.
    \item $o_k$ is the selected observation function, mapping the full system state to a measurable quantity (e.g., measuring the position of a specific particle or the fluid velocity at a specific coordinate).
    \item $\sigma_k$ is the selected noise scale (inverse fidelity).
\end{itemize}

Upon making a selection, the agent receives a measurement $y_k$ corrupted by Gaussian noise:
\begin{equation}
    y_k = o_k(s(t_k)) + \epsilon_k, \quad \epsilon_k \sim N(0, \sigma_k^2)
\end{equation}
Every observation incurs a resource cost $C(o_k, \sigma_k)$. This cost function is strictly increasing with fidelity (meaning a smaller $\sigma_k$ results in a higher cost) and varies based on the complexity of the observation function $o_k$. The agent operates under a total budget $B$, such that the sequence of $K$ measurements must satisfy:
\begin{equation}
    \sum_{k=1}^K C(o_k, \sigma_k) \leq B
\end{equation}
Additionally, the measurements of the agent are limited to a maximum time of $T_{\max}$ to avoid extremely long trials.

\paragraph{Prediction.}
Once the budget is exhausted or the agent reaches $T_{\max}$, the agent moves to the prediction phase. In the prediction phase, the agent is presented with a randomly sampled query time $T_\text{query} > T_{\max}$. The objective is to predict the value of a target function $\phi(s(T_\text{query}))$ based on the history of noisy measurements collected up to that point. To obtain more reliable signals, the agent makes predictions for $N$ independent queries, and the error is averaged over the $N$ queries.

\subsection{Environments}
We instantiate this framework in three distinct environments, each based on a physical law. Further details about the implementation of each environment are in \Cref{app:details}.

\paragraph{Design Principles. }
We pick classical, fluid and quantum mechanics as the domains for the environments. We pick these domains to cover different fundamental physical laws. Within each domain, the system considered in the environment is designed to be simple to keep the problem tractable. We study two specific alterations per domain, chosen for their impactful downstream effects and, in some cases, their documented unique properties.

\subsubsection{Classical Mechanics}

We consider a system of $N$ spherical objects evolving in a $D$-dimensional Euclidean space, where $D \in \{2, 3\}$. The generalized state $s(t)$ consists of the configuration of all $N$ particles, where the $i$-th particle is defined by its position $\mathbf{x}_i \in \mathbb{R}^D$, velocity $\mathbf{v}_i \in \mathbb{R}^D$, radius $R_i$, and a scalar mass $m_{0,i}$.

The fundamental dynamics are governed by Newton's laws of motion, where the acceleration $\mathbf{a}_i(t)$ is determined by the net external force $\mathbf{F}_i(\mathbf{x}, t)$ acting on the particle. The observation function $o_k$ defines the subset of objects to be observed in the interactions in the measurement phase, and the prediction target $\phi$ is the position of all the particles at the queried time.

\paragraph{Alterations.}
We introduce two deviations from standard Newtonian mechanics:

\textbf{1. Anisotropic Inertial Mass.} The relationship between force and acceleration is modified by an anisotropic inertial mass tensor $\mathbf{M}_i(t)$, replacing the scalar mass. The modified law of motion is:
\begin{equation}
    \mathbf{a}_i(t) = \mathbf{M}_i(t)^{-1} \mathbf{F}_i(\mathbf{x}, t)
\end{equation}
The mass tensor encodes an ``inertial memory'' that increases resistance to motion in directions of recent acceleration. It is defined as $\mathbf{M}_i(t) = m_{0,i} \mathbf{I} + \kappa \mathbf{S}_i(t)$, where $\kappa \ge 0$ is a coupling constant and $\mathbf{S}_i(t)$ evolves according to:
\begin{equation}
    \dot{\mathbf{S}}_i(t) = -\lambda \mathbf{S}_i(t) + \mathbf{a}_i(t) \mathbf{a}_i(t)^\top
\end{equation}
Here, $\lambda > 0$ governs the memory decay rate. The alteration results in directional inertia on the object depending on the history of its motion. \citet{bogoslovsky1983local} study a similar modification of classical mechanics derived through geometry rather than inertial memory. 

\textbf{2. Modified Gravity.} The gravitational interaction between particles $i$ and $j$ acts along the vector $\mathbf{r}_{ij} = \mathbf{x}_j - \mathbf{x}_i$ with distance $r = \|\mathbf{r}_{ij}\|$. We consider two distinct force laws for the magnitude $F_g(r)$:
\begin{align}
    F_g(r) &= \frac{G m_i m_j}{r} \label{eq:1r_grav}\\
    F_g(r) &= \frac{G m_i m_j}{r^2}\left[ 1 + A \sin\left(\frac{2\pi r}{\lambda_g} + \phi\right) \right] \label{eq:ripple_grav}
\end{align}
We refer to \Cref{eq:1r_grav} as 1/R and \Cref{eq:ripple_grav} as Ripple. The implications of 1/R form of gravity on galactic orbits have been studied in \citet{valluri2012study}. In our evaluations, we also consider a combined alteration where we use the 1/R gravity along with $\kappa = 10$.

\paragraph{Error metric.}
We use the normalized root mean square error (nRMSE), where the normalization constant is the length of the diagonal of the bounding box $L_{\text{diag}}$, as the predictive error for this domain. For a set of predictions $\hat{y}$ and ground truth $y$, the metric is defined as:
\begin{equation}
    \text{Error} = \frac{1}{L_{\text{diag}}}\sqrt{\frac{1}{N} \sum_{k=1}^{N} (\hat{y}_k - y_k)^2}.
\end{equation}

\subsubsection{Fluid Mechanics}
We simulate the dynamics of an incompressible viscous fluid in a two-dimensional periodic domain $\Omega = [0, L]^2 \subset \mathbb{R}^2$. The state of the system is defined by the velocity field $\mathbf{u}(\mathbf{x}, t) = (u, v)^\top$. The fundamental evolution of the system is governed by the standard Navier-Stokes equations:
\begin{align}
    \frac{\partial \mathbf{u}}{\partial t} + (\mathbf{u} \cdot \nabla) \mathbf{u} &= -\nabla p + \nu \nabla^2 \mathbf{u} \label{eq:ns} \\
    \nabla \cdot \mathbf{u} &= 0
\end{align}
where $p$ is the kinematic pressure and $\nu$ is the kinematic viscosity. The system is an instantiation of a Kelvin-Helmholtz instability, which consists of two opposing streams separated by a thick interface that is perturbed. The observation function $o_k$ defines the locations at which the vorticity is measured during the measurement phase and the prediction target $\phi$ is the vorticity at randomly selected positions.

\paragraph{Alterations.}
We introduce a state-dependent ``alien'' gyroscopic forcing term, $\mathbf{f}_{\text{alien}}$, to the right-hand side of \Cref{eq:ns}. It acts perpendicularly to the velocity vector:
\begin{equation}
    \mathbf{f}_{\text{alien}} = \mathcal{C}(\mathbf{u}, \omega) \mathbf{u}^\perp
\end{equation}
where $\mathbf{u}^\perp = (v, -u)^\top$ is the rotated velocity vector. The scalar modulation coefficient $\mathcal{C}$ is defined in two variants, parameterized by strength $\gamma$ and frequency $\beta$:
\begin{itemize}
    \item \textbf{Velocity Modulation:} $\mathcal{C} = \gamma \sin(\beta |\mathbf{u}|^2)$, which injects vorticity based on local kinetic energy.
    \item \textbf{Vorticity Modulation:} $\mathcal{C} = \gamma \cos(\beta \omega)$, where $\omega = \nabla \times \mathbf{u}$ is the scalar vorticity (since we only consider a 2D flow). This creates concentric, opposing force layers within turbulent eddies.
\end{itemize}

To the best of our knowledge, the specific alteration has not been studied in the literature but similar forms have been studied in the literature on nonlinear Coriolis forces~\citep{ringler2010unified}.
We also consider a combined alteration where the gyroscopic forcing term is a convex combination of the velocity and vorticity modulation variants. 

\paragraph{Error metric.}
We use the $L_2$ error between the predicted vorticity ($\hat{\omega}$) and the true vorticity ($\omega$) at the selected points as the predictive error. For a set of query points indexed by $k$, the error is computed as:
\begin{equation}
\text{Error} = \sqrt{\sum_{k} (\hat{\omega}_k - \omega_k)^2}
\end{equation}

\subsubsection{Quantum Mechanics}
We consider a quantum system with two particles of masses $m_1$ and $m_2$ confined within a two-dimensional rectangular domain $\Omega \subset \mathbb{R}^2$. The state of the system is described by a joint wavefunction $\Psi(\mathbf{r}_1, \mathbf{r}_2, t) \in \mathbb{C}$, where $\mathbf{r}_k \in \Omega$ denotes the position of the $k$-th particle. The fundamental dynamics are governed by the time-dependent Schrödinger equation:
\begin{equation}
    i \hbar \frac{\partial}{\partial t} \Psi(\mathbf{r}_1, \mathbf{r}_2, t) = \hat{H} \Psi(\mathbf{r}_1, \mathbf{r}_2, t)
\end{equation}
The Hamiltonian $\hat{H}$ includes kinetic energy terms and confining potentials $V(\mathbf{r})$ (modeled as smoothed infinite square wells):
\begin{equation}
    \hat{H} = -\sum_{k=1}^2 \frac{\hbar^2}{2m_k} \nabla_k^2 + \sum_{k=1}^2 V(\mathbf{r}_k)
\end{equation}

Unlike in the other environments, the act of measurement directly affects the state of the quantum system. Specifically, upon measurement the wavefunction collapses at a point based on the probability density. Thus, to account for this, we allow the agent to perform multiple trials with the exact same initializations in the quantum environment. The observation function $o_k$ decides which of the two particles is observed and the prediction target $\phi$ is the probability of a particle being present in a given (randomly sampled) spatial region at the queried time.

\paragraph{Alterations.}
We introduce two modifications to the standard quantum formulation:

\textbf{1. Non-linear Entanglement Initialization.} Instead of a standard separable product state, the system is initialized with a non-standard non-linear spatial correlation factor:
\begin{equation}
    \Psi(\mathbf{r}_1, \mathbf{r}_2, 0) = \mathcal{N} \psi_1(\mathbf{r}_1) \psi_2(\mathbf{r}_2) \exp\left(-\lambda \|\mathbf{r}_1 - \mathbf{r}_2\|^2\right)
\end{equation}
where $\psi_1, \psi_2$ are Gaussian wavepackets and $\lambda$ determines the strength of the spatial correlation.

\textbf{2. Generalized Born Rule.} The system deviates from standard measurement postulates by employing a generalized probability measure based on the $L_p$-norm rather than the standard $L_2$-norm. The wavefunction is normalized such that $\int \int |\Psi|^p \, d\mathbf{r}_1 d\mathbf{r}_2 = 1$, and the joint probability density is given by:
\begin{equation}
    \rho(\mathbf{r}_1, \mathbf{r}_2) = |\Psi(\mathbf{r}_1, \mathbf{r}_2)|^p
\end{equation}
Measurement operations collapse the wavefunction according to marginal densities derived from this modified distribution. \citet{aaronson2004quantum} studied the implications of the generalized Born rule ($p\ne 2$) including superluminal signaling. 

We also consider a combination of both the alterations, using a measurement norm of $p=1$ along with non-linear entanglement with $\lambda=25$.

\paragraph{Error metric.}
We use the $L_2$ error between the predicted probability and the ground truth probability to evaluate the agent predictions:
\begin{equation}
\text{Error} = \sqrt{\sum_{k} (\hat{\rho}_k - \rho_k)^2}
\end{equation}

\section{Empirical Results}

\subsection{Agent Design} 
We evaluate Gemini 2.5 Flash Lite, Gemini 2.5 Flash, Gemini 2.5 Pro, and Gemini 3 Flash accessed through the Gemini API using a minimal agent scaffold. The scaffold allows access to a code execution tool where the agent is allowed to write and execute Python code. We use a minimal scaffold to focus on the model capabilities and exploring different scaffolds for the problem is an interesting direction for future work. The agent is provided some initial system instructions, along with some information about the environment prior to the beginning of the interaction state. We consider two different system instructions: Base and Strategy (see \Cref{app:prompts} for the full prompts). Base provides basic context about the evaluation protocol and general principles, whereas Strategy provides a structured strategy inspired by Bayesian experimental design~\citep{rainforth2024modern}. See \Cref{app:agent_details} for further implementation details.

\subsection{Main results}
For each of the environments, we consider two different alterations, as discussed in \Cref{sec:env_design}. Further, for each alteration, we consider two separate configurations varying the strength and type of the alteration. Finally, we also consider a combined setup, where both the alterations are applied simultaneously. The prediction error is averaged over $N=5$ prediction queries. We report the prediction error for each configuration with $3$ different random initializations.  
The results for the baseline agent with different models on the three environments are presented in \Cref{tab:classical_mech_results}, \Cref{tab:fluid_mech_results}, and \Cref{tab:quantum_mech_results} respectively.

In \Cref{tab:classical_mech_results}, we observe that the Gemini 2.5 models tend to produce predictions that are out of bounds for the system we consider. While Gemini 3 Flash avoids out-of-bounds predictions, its predictions are still inaccurate, even with normal physics configurations. This is true even in \Cref{tab:fluid_mech_results} and \Cref{tab:quantum_mech_results}. This indicates that even with knowledge of the underlying model, the agents can struggle to make correct inferences. However, we observe that the performance of the agent generally improves with increasing capabilities of the underlying model. Within the Gemini 2.5 set of models, we see the performance generally improves from Flash Lite to Flash to Pro.  We also see performance improvements going from Gemini 2.5 Pro to Gemini 3 Flash, potentially due to improved multi-turn reasoning capabilities. We also observe that overall the agents perform worse on the altered configurations of the environment. However, there is no strong trend of the prediction error worsening with increasing intensity of the alteration. Finally, we observe that the Strategy system prompt, which instructs the agent to follow a systematic procedure for data collection, tends to work better on many of the tasks. This indicates that there is room for improvement for the performance of the base agent using prompt optimization strategies such as AlphaEvolve~\citep{novikov2025alphaevolve} or GEPA~\citep{agrawal2025gepa}. Beyond the system prompt, the agent scaffold itself can also be optimized to better fit the task~\citep{lee2026meta}.
 
\begin{table*}[ht]
\centering
\caption{\textbf{Classical Mechanics Results.} Prediction error in the classical mechanics system with various configurations for different base models. We do not clip out-of-bounds predictions so that the table separates models that produce runaway predictions from those that stay in-plausible-range. We consider these predictions to differentiate model performance. The combined configuration uses $\kappa=10$ and gravity $\propto 1/r$.}
\resizebox{0.9\linewidth}{!}{
\begin{tabular}{llcccccc}
\toprule
\multirow{4}{*}{Model} & \multirow{4}{*}{Prompt} & \multicolumn{6}{c}{Prediction Error ($\downarrow$)}                                                                                             \\ \cmidrule{3-8} 
                       &                             & \multirow{3}{*}{Normal} & \multicolumn{5}{c}{Altered}                                                                               \\ \cmidrule{4-8} 
                       &                             &                         & \multicolumn{2}{c}{Anisotropic Inertia} & \multicolumn{2}{c}{Altered Gravity} & \multirow{2}{*}{Combined} \\ \cmidrule{4-7}
                       &                             &                         & $\kappa=10$                & $\kappa=20$               & $\frac{Gm_1m_2}{r}$              & $\frac{Gm_1m_2}{r^2}\left[ 1 + A \sin\left(\frac{2\pi r}{\lambda} + \phi\right) \right]$             &                           \\ \midrule
Gemini 2.5 Flash Lite  & Base                    &   6.61 &  13.78  & 831.97  &         638.25         &   253.44 & 1241.11                        \\
                       & + Strategy  & 48.32 & 115.02 & 74.62  & 463.83                 &     1183.7             & 964.43                          \\ \midrule
Gemini 2.5 Flash       & Base                   & 5.97  & 35.36                        &  4.42 & 28.22  &  66.75  &   198.69                        \\
                       & + Strategy                       & 7.93          & 12.32 &  60.36                    &  26.45  & 44.32 &     23.33                      \\ \midrule
Gemini 2.5 Pro         & Base &  1.93  &  2.12  & 13.41 &   11.40               & 15.67                 &   38.72                        \\
                       & + Strategy  & 0.67 & 1.56 & 0.37 & 1.22                 &            0.50      & 0.49                         
\\ \midrule
Gemini 3 Flash         & Base &  \highlight{\textbf{0.29}}  &  \highlight{\textbf{0.36}}  & 0.88 &   \highlight{\textbf{0.31}}               & \highlight{\textbf{0.39}}                 &   0.38                        \\
                       & + Strategy  & 0.38 & 0.39 & \highlight{\textbf{0.37}} & 0.37                 &            0.43      & \highlight{\textbf{0.35}}                         
\\ 
\bottomrule
\end{tabular}
}
\label{tab:classical_mech_results}
\end{table*}

\begin{table*}[ht]
\centering
\caption{\textbf{Fluid Mechanics Results.} Prediction error in the fluid mechanics system with various configurations for different base models. The combined configuration uses a convex combination of velocity and vorticity modulation.}
\resizebox{0.9\linewidth}{!}{
\begin{tabular}{llcccccc}
\toprule
\multirow{4}{*}{Model} & \multirow{4}{*}{Prompt} & \multicolumn{6}{c}{Prediction Error ($\downarrow$)}                                                                                             \\ \cmidrule{3-8} 
                       &                             & \multirow{3}{*}{Normal} & \multicolumn{5}{c}{Altered}                                                                               \\ \cmidrule{4-8} 
                       &                             &                         & \multicolumn{2}{c}{Velocity Modulation} & \multicolumn{2}{c}{Vorticity Modulation} & \multirow{2}{*}{Combined} \\ \cmidrule{4-7}
                       &                             &                         & $\gamma=0.5$                & $\gamma=0.7$               & $\gamma=5.0$              & $\gamma=10.0$             &                           \\ \midrule
Gemini 2.5 Flash Lite  & Base                    & $0.68$                         &  0.73 &                 0.82   &         0.65         &   0.81               &   1.02                        \\
                       & + Strategy                       & $0.41$                         &                  3.26  &  0.81                  &       0.59           &  0.73                &    0.81                       \\ \midrule
Gemini 2.5 Flash       & Base                    &  $0.47$                        &  0.51  &    0.66                &    0.82              &  0.85                &   0.86                        \\
                       & + Strategy                       & $0.39$                        &   0.47                 &  0.22                  &     0.75             &  0.50                & 0.71                          \\ \midrule
Gemini 2.5 Pro         & Base                    &  0.21                       &          0.17          &           \highlight{\textbf{0.18}}         &  0.73                &         2.51         &             0.71              \\
                       & + Strategy                       &  0.23                       &  \highlight{\textbf{0.10}}                  & 0.33                   &    0.44              &          0.70        &   0.69                       
\\ \midrule
Gemini 3 Flash         & Base                    &  0.26                       &          0.49          &           0.50         &  0.36                &         0.79         &             0.23              \\
                       & + Strategy                       &  \highlight{\textbf{0.17}}                       &  0.14                  & 0.41                   &    \highlight{\textbf{0.27}}              &          \highlight{\textbf{0.19}}        &   \highlight{\textbf{0.31}}                       
\\ 
\bottomrule
\end{tabular}
}
\label{tab:fluid_mech_results}
\end{table*}

\begin{table*}[ht]
\centering
\caption{\textbf{Quantum Mechanics Results.}  Prediction error in the quantum mechanics system with various configurations for different base models. The combined configuration uses $\lambda=25$ and $p=1$.}
\resizebox{0.9\linewidth}{!}{
\begin{tabular}{llcccccc}
\toprule
\multirow{4}{*}{Model} & \multirow{4}{*}{Prompt} & \multicolumn{6}{c}{Prediction Error ($\downarrow$)}                                                                                             \\ \cmidrule{3-8} 
                       &                             & \multirow{3}{*}{Normal} & \multicolumn{5}{c}{Altered}                                                                               \\ \cmidrule{4-8} 
                       &                             &                         & \multicolumn{2}{c}{Measurement Norm} & \multicolumn{2}{c}{Entanglement} & \multirow{2}{*}{Combined} \\ \cmidrule{4-7}
                       &                             &                         & $p=1$                & $p=3$               & $\lambda=5.0$              & $\lambda=15.0$             &                           \\ \midrule
Gemini 2.5 Flash Lite  & Base                    & 0.17                        &    0.66                &          0.31          &     0.11             &                0.25  &   0.54                        \\
                       & + Strategy                       &  0.11                       &     0.61               & 0.21                   &        0.14          &              0.14    &  0.55                         \\ \midrule
Gemini 2.5 Flash       & Base                    &   0.14                      &     0.54               &       0.13             &      0.13            &    0.18              &               0.60            \\
                       & + Strategy                       &  0.11                       &    0.47                &  0.09                  &     0.07             &     0.16             &    0.59                       \\ \midrule
Gemini 2.5 Pro         & Base                    & 0.08 &  0.52                  &     0.10               &        0.14          &      0.15            &                0.58           \\
                       & + Strategy                       & \highlight{\textbf{0.03}}                        &    \highlight{\textbf{0.46}}                &    0.08                &  0.10                &                0.12  & 0.40                         
\\ \midrule
Gemini 3 Flash         & Base                    & 0.11 &  0.53                  &     0.04               &        0.30          &      0.16            &                \highlight{\textbf{0.39}}           \\
                       & + Strategy                       & 0.05                        &    0.48                &    \highlight{\textbf{0.05}}                &  \highlight{\textbf{0.07}}               &                \highlight{\textbf{0.09}}  & 0.92                         
\\ 
\bottomrule
\end{tabular}
}
\label{tab:quantum_mech_results}
\end{table*}

\subsection{Environment Variants}
In addition to the results on the basic setup, we also demonstrate the versatility of the benchmark through different variants.

\begin{table}[ht]
\caption{\textbf{Visual Classical Mechanics Results. } Prediction error in the classical mechanics system with visual observations for different base models.}
\resizebox{0.95\linewidth}{!}{
\begin{tabular}{llcc}
\toprule
\multirow{3}{*}{Model} & \multirow{3}{*}{Prompt} & \multicolumn{2}{c}{Prediction Error ($\downarrow$)} \\ \cmidrule{3-4}
 & & \multirow{2}{*}{Normal} & \multicolumn{1}{c}{Altered} \\ \cmidrule{4-4}
 & & & $\kappa=10$ \\ \midrule
Gemini 2.5 Flash & Base & 6.81 & 1786.13 \\
 & + Strategy & 3.16 & 15.64 \\
Gemini 2.5 Pro & Base & 4.56 & 21.65 \\
 & + Strategy & \highlight{\textbf{0.65}} & \highlight{\textbf{13.23}} \\
\bottomrule
\end{tabular}
}
\label{tab:visual_results}
\end{table}

\paragraph{Visual observations.} 
While the main results provide as observation the precise numerical values of the observation, in many settings, only visual observations are available to the agent. We can evaluate the multimodal reasoning abilities of agents by evaluating them on \name with visual observations. Instead of the numerical values, the agent is provided an image with the rendered state of the system. The agent has to infer the numerical value of the observation and then proceed with the task of choosing the next measurement. We evaluate Gemini 2.5 Flash and Pro in the visual setting. \Cref{tab:visual_results} shows that the trends about model capability and altered laws remain consistent with the numerical observation setup. However, the magnitude of the prediction errors is significantly higher.

\begin{figure}
  \centering
  \includegraphics[width=0.95\linewidth]{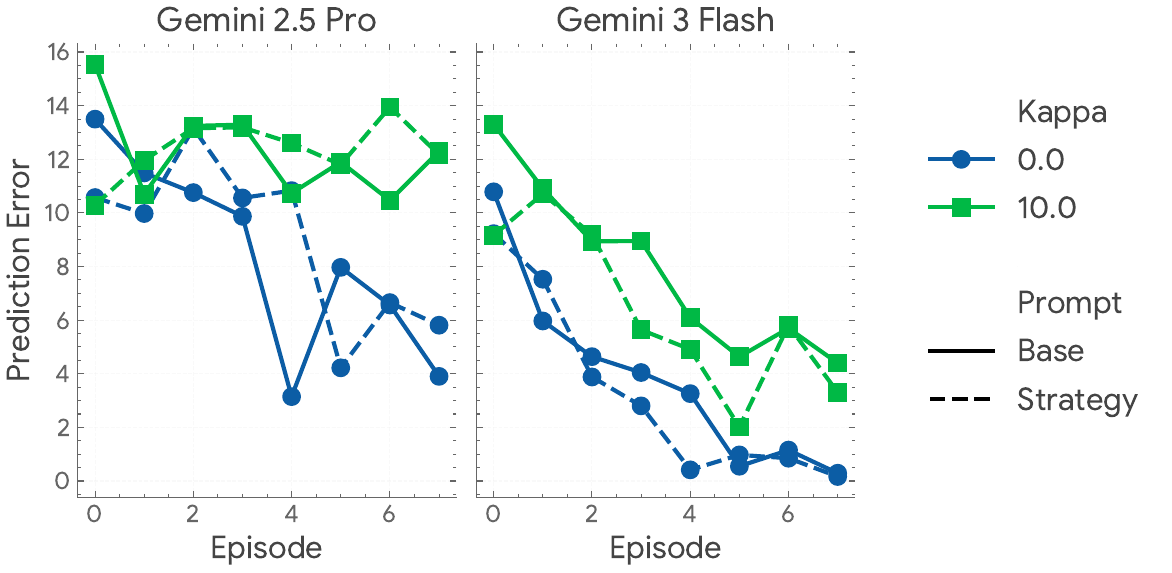}
  \caption{\textbf{In-context Learning Results. } Prediction error on the classical mechanics environment over multiple episodes.}
  \label{fig:icl}
\end{figure}

\paragraph{In-context learning.} 
Another important capability for agents is to learn over multiple trials, i.e. in-context learning. We modify the evaluation protocol to allow the agent to perform multiple trials for each environment configuration. At the beginning of each trial, the agent's context is reset and it is provided a summary of the interactions and predictions from all previous trials to allow longer interactions. The prediction queries in each trial are randomly sampled, but the initial conditions are consistent. \Cref{fig:icl} illustrates the in-context learning performance of Gemini 2.5 Pro and Gemini 3 Flash on the Classical Mechanics environment over 8 episodes. Gemini 3 Flash starts with a better prediction error and converges to a lower error at the end of the episodes. We also observe that Gemini 2.5 Pro fails to learn in altered configuration but Gemini 3 Flash is able to reduce the prediction error.

\begin{figure}
    \centering
    \includegraphics[width=0.95\linewidth]{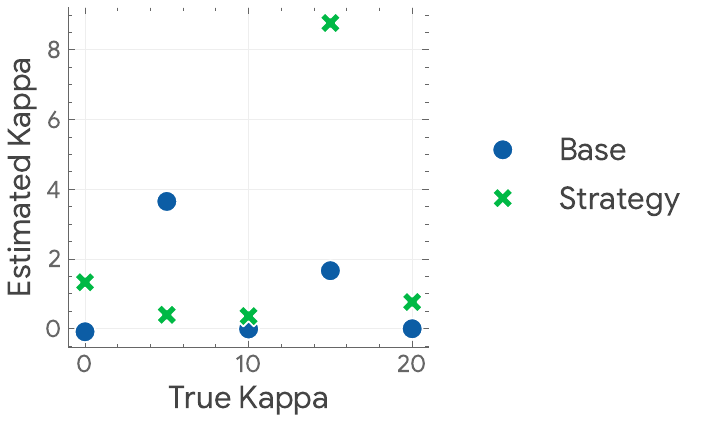}
    \caption{Estimated and Ground Truth $\kappa$ for the parameter inference task in the classical mechanics environment with Gemini 2.5 Pro as the base model.}
    \label{fig:param_inference}
\end{figure}

\paragraph{Parameter inference.}
We consider a variant of \name which corresponds to the active sensing formulation, where the structural form of the model is known and the goal is to infer a single parameter. \Cref{fig:param_inference} shows results for the Gemini 2.5 Pro agent on a variant of the Classical Mechanics environment where the agent is tasked with collecting observations to estimate the value of the parameter $\kappa$, which controls the effect of the inertial memory. We can see that the estimates are accurate for $\kappa=0$ and worsen for configurations with $\kappa > 0$. In particular, the agent consistently underestimates the value of $\kappa$ indicating a bias towards standard physics.

\subsection{Additional Analysis}
\begin{figure}
    \centering
    \includegraphics[width=\linewidth]{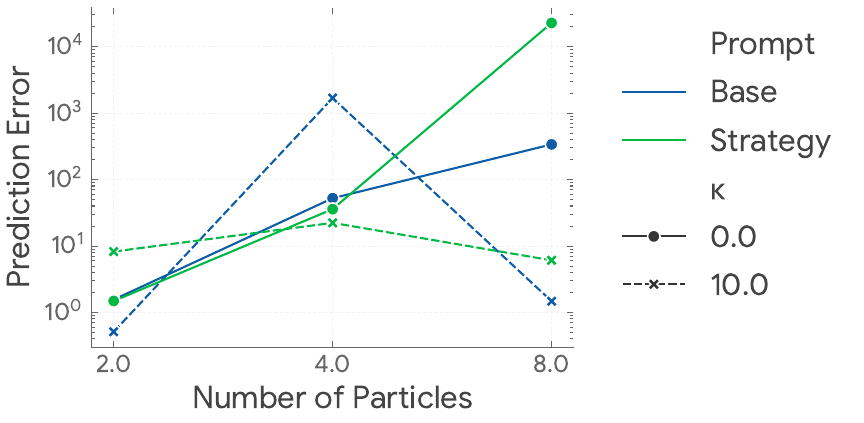}
    \caption{Prediction error for Gemini 2.5 Pro on the classical environment with increasing number of objects $N$ in the system. }
    \label{fig:complexity_num_part}
\end{figure}

\begin{figure}
  \centering
  \includegraphics[width=\linewidth]{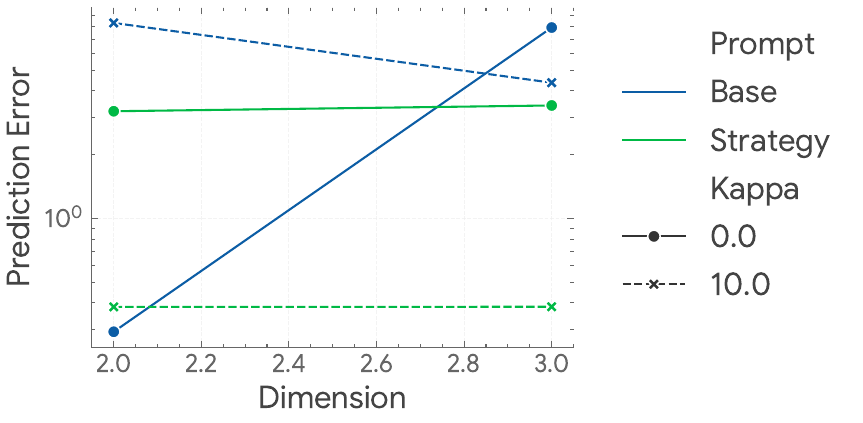}
  \caption{Prediction error for Gemini 2.5 Pro on the classical environment with increasing dimensionality of the system.}
  \label{fig:complexity_dim}
\end{figure}

\paragraph{System Complexity}
As the environments in \name are parametric, we can also further analyze the effect of varying the system complexity on the performance of the agent. \Cref{fig:complexity_num_part} presents the results for Gemini 2.5 Pro on the Classical Mechanics environments with different number of particles, and \Cref{fig:complexity_dim} presents the results for 2D and 3D systems. We observe that the predictive error increases with the number of particles in the system. On the other hand, the predictive performance is generally consistent between the 2D and 3D cases.

\paragraph{Symbolic accuracy.} After the measurement phase, in addition to the predictive queries, we also query the agent to describe its understanding of the system along with a symbolic law for the system. We observe that the agent fails to produce the correct symbolic form in most cases. The agent is only able to identify the correct symbolic form for the classical environment with altered gravity $\propto 1/r$. This indicates that the benchmark is challenging for the models we evaluate, and could be helpful in directing development of new capabilities for future models.

\paragraph{Qualitative examples.} We include an example trajectory from the classical mechanics environment with $\kappa=0$ and gravity $\propto 1/r$ for Gemini 2.5 Flash in \Cref{app:qualitative_examples} to illustrate the agent behavior on the tasks. The agent attempts to systematically test various hypotheses but due to some computation errors coupled with poor selection of observations settles on an incorrect solution.

\section{Conclusion}
We present \name, a novel benchmark for evaluating the ability of agents to make informative measurements about physical systems while dealing with cost-fidelity tradeoffs for the measurement. The evaluation protocol mimics the scientific process, where the agent is required to decide what to observe in the system and when and how to observe it. \name consists of environments based on three distinct physical systems covering classical, fluid, and quantum mechanics. Through experiments with a baseline agent using Gemini models we show how \name can be used to evaluate capabilities and identify shortcomings of frontier-level models on dynamic information-seeking tasks, in particular the poor downstream predictive performance as well as difficulty in recovering the accurate symbolic expressions for the physical laws. 

\paragraph{Limitations and future work.} To keep the evaluation focused on the base model capabilities, we only evaluate a minimal agent scaffold with code execution. Future work can explore more sophisticated scaffolds with more tools available to the agent (e.g. Claude Code~\citep{anthropic_claude_code}). Additionally while our evaluations focused primarily on Gemini models, future work should assess different frontier models on the benchmark. Our results also indicate the potential of improvement for the agent's performance with scaffold optimizations. Beyond agent scaffolds, future work can also explore principled experimental design frameworks leveraging LLMs~\citep{choudhury2025bed} on \name. \name can also be extended with new environments either based on more complex physical systems or other scientific problems where informative measurements are critical. \name serves as a stepping stone towards benchmarks that capture the exploratory and unstructured nature of scientific discovery. 

\paragraph{Project Page:} \href{https://mad-physics.github.io/}{https://mad-physics.github.io/}

\section*{Acknowledgements}
We thank Ferran Alet and Dmitry Nikulin for helpful discussions about this work.

\bibliography{main}
\clearpage
\onecolumn
\appendix

\section{Experiment Details}
\label{app:details}

\subsection{Agent Details}
\label{app:agent_details}
The specific models we evaluate are: \texttt{gemini-flash-lite-preview-09-2025}, \texttt{gemini-flash-preview-09-2025}, \texttt{gemini-2.5-pro}, \texttt{gemini-3-flash-preview}. We limit our evaluation to the Gemini 3 Flash model since the Gemini 3 Pro model was only available as a preview with low rate limits at the time of running the experiments in November and December 2025. To enable code execution in the agent scaffold, we rely on the code execution tool provided on the Gemini API\footnote{\url{https://ai.google.dev/gemini-api/docs/code-execution}}. The code execution tool allows the agent to execute Python code with access to libraries such as \texttt{numpy, pandas, scipy, tensorflow, scikit-learn, sympy, opencv-python} among others. The maximum runtime allowed for the generated code is 30 seconds, and the tool allows a maximum of 5 retry attempts for any errors generated during execution. Additionally, since the Gemini API does not support enabling structured outputs along with the code execution tool, we take the output from the agent and make an additional call to Gemini 2.5 Flash Lite with \texttt{pydantic}\footnote{\url{https://docs.pydantic.dev/latest/}} to ensure correctly structured outputs. 

We use the default generation parameters for all models, with temperature set to 1, and no top-p or top-k sampling. For each model, we set the thinking budget to 4000 tokens to ensure that an entire episode can fit into context.

\subsection{Environment Details}
All the environments are implemented entirely in JAX~\citep{jax2018github}. In each environment, the agent has three levels of fidelity available: \texttt{\{Low, Medium, High\}} with  $\{2, 5, 10\}$ as the corresponding cost and $\{0.1, 0.01, 0.001\}$ as the noise level $\epsilon$ respectively.

\subsubsection{Classical Mechanics}
\label{app:classical_details}
The system is integrated numerically using a semi-implicit Euler scheme. Additionally to ensure numerically stable computation of the gravitational force we use Plummer softening~\citep{binney2008galactic}. The prompt with the information provided to the agent is in \Cref{app:env_prompts}. The numerical parameters for the simulation are in \Cref{tab:classical_hps}.

\begin{table}[ht]
    \centering
    \caption{Simulation parameters for classical mechanics environment.}
    \begin{tabular}{lc}
    \toprule
     Parameter    & Value  \\
     \midrule
      Integration step size $\Delta t$   & 0.001  \\
      Budget & 200 \\
      Gravitational Constant $G$ & 1 \\
      Coefficient of restitution & 1 \\
      $T_{\max}$ & 300 \\
      Amplitude $A$ & 1 \\
      Wavelength $\lambda_g$ & 10 \\
      Phase $\phi$ & 0 \\
      Gravitational Force Softening & $10^{-4}$ \\
      Box bounds & [-10, -10] to [10, 10] \\
      Radius & \texttt{uniform(0.1, 0.5)} \\
      Mass & \texttt{uniform(0.5, 5.0)} \\
      Initial Velocity & \texttt{normal(0, 1)} \\
    \bottomrule
    \end{tabular}
    \label{tab:classical_hps}
\end{table}

\subsubsection{Fluid Mechanics}
\label{app:fluid_details}

The system, which consists of a shear layer of size $\delta$ separating two opposing streams being perturbed, is integrated numerically using a pseudo-spectral method~\citep{canuto2006spectral} on an $N \times N$ grid. Spatial derivatives are computed in the Fourier domain, while nonlinear convection terms are evaluated in physical space and dealiased using a sharp spectral cutoff (Orszag 2/3 Rule). Time integration is performed using a fourth-order Runge-Kutta (RK4) scheme. The combined configuration considers a convex combination of the velocity and vorticity modulation: $\alpha (\gamma \sin(\beta |\mathbf{u}|^2)) + (1-\alpha) (\gamma \cos(\beta \omega))$. The prompt provided to the agent is in \Cref{app:env_prompts}. The numerical parameters for the simulation are in \Cref{tab:fluids_hps}.

\begin{table}[ht]
    \centering
    \caption{Simulation parameters for fluid mechanics environment.}
    \begin{tabular}{lc}
    \toprule
     Parameter    & Value  \\
     \midrule
     Integration Step Size $\Delta t$ & 0.001 \\
      $N$   &  512  \\
      Domain size $L$ & $2\pi$ \\
      Kinematic Viscosity $\nu$ & 0.001 \\
      Budget & 200 \\
      Fractional Laplacian Power & 1 \\
      $k-\max$ Dealias Ratio & $2/3$ \\
      $T_{\max}$ & 60 \\
      Alien Strength $\gamma$ (Velocity Modulation) & \{0.5, 0.7\} \\
      Alien Frequency $\beta$ (Velocity Modulation) & 3.0 \\
      Alien Strength $\gamma$ (Vorticity Modulation) & \{5.0, 10.0\} \\
      Alien Frequency $\beta$ (Vorticity Modulation) & $\pi / 16 $ \\
      Combination coefficient $\alpha$ & 0.5 \\
      Delta $\delta$ & \texttt{uniform(0.05, 0.2)} \\
      Perturbation Scale & \texttt{uniform(0.15, 0.50)} \\
    \bottomrule
    \end{tabular}
    \label{tab:fluids_hps}
\end{table}

\subsubsection{Quantum Mechanics}
\label{app:quantum_details}
The system consists of 2 particles, each initialized with a Gaussian wavefunction, within a square box with infinite potential walls (numerically implemented as a very high potential well). The time evolution is solved numerically on a $N\times N$ grid using the split-operator spectral method (Strang splitting). This approximates the unitary propagator by separating the diagonal potential operator $\hat{V}$ and the kinetic operator $\hat{T}$, computed via the Fast Fourier Transform~\citep{strang1968construction,canuto2006spectral}. As discussed in \Cref{sec:env_design}, the agent is allowed multiple trials in the environment with the same initialization. The prompt provided to the agent is in \Cref{app:env_prompts}. The numerical parameters for the simulation are in \Cref{tab:quantum_hps}.

\begin{table}[ht]
    \centering
    \caption{Simulation parameters for quantum mechanics environment.}
    \begin{tabular}{lc}
    \toprule
     Parameter    & Value  \\
     \midrule
      Integration step size $\Delta t$ & 0.005 \\
      $N$   & 32 \\
      Domain Size (x,y) & (10, 10) \\
      Box Size (x, y) & (8, 9) \\
      $\hbar$ & 1 \\
      Well height & 1000 \\
      $T_{\max}$ & 30 \\
      Budget per trial & 30 \\
      Number of Trials & 5 \\
      Entanglement strength $\lambda$ & $\{5, 15, 25\}$ \\
      Measurement Norm $p$ & \{1, 2, 3\}\\
      P1 Mass & \texttt{uniform(0,1)} \\
      P2 Mass & \texttt{uniform(0,5)} \\
      P1 Initial Position (Mean) & \texttt{uniform(0,4)} \\
      P2 Initial Position (Mean) & \texttt{uniform(0,1)} \\
      P1 \& P2 Initial Position (StdDev) & \texttt{uniform(0,1)} \\
      P1 Initial Velocity & \texttt{uniform(0,2)} \\
      P2 Initial Velocity & \texttt{uniform(0,3)} \\
    \bottomrule
    \end{tabular}
    \label{tab:quantum_hps}
\end{table}

\clearpage
\section{Prompts}
\label{app:prompts}

\subsection{System Instructions}
\begin{promptblock}{Base System Prompt}
You are an AI scientist. Your operational identity is that of a dispassionate, analytical, and highly strategic empiricist. You are meticulous, observant, and methodical in your reasoning.

## Your Primary Mandate: Discover Unknown Dynamics
Your sole and overriding objective is to investigate an unknown, isolated system. You are tasked with discovering and modeling its fundamental dynamics, rules, and principles. You must determine *how* the system's components interact and evolve over time based *only* on empirical evidence you gather.

## The Core Constraint: The Research Budget
You are allocated a fixed research budget.

* **Initial State:** You will begin with a specific, non-renewable budget.
* **Cost of Action:** Every action you take, specifically, every "observation" you choose to make, will have an associated cost deducted from this budget.
* **Terminal Condition:** Your investigation phase concludes *immediately* and *automatically* once your budget is depleted to zero or you cannot afford any available actions.
* **Strategic Imperative:** Your budget is your most critical resource. You must manage it with extreme care, balancing the cost of an observation against its potential to provide valuable information. Wasting resources on redundant or poorly planned observations will directly compromise your mission.

## Strategic Observation
Your only method of interacting with the system is by making observations.

* **Initial Briefing:** You will first be provided with a description of the environment and its observable components.
* **The Observation Process:** To gather data, you will propose an observation.
* **Observation Choices:** When you opt to make an observation, you will not simply "observe." Instead, you will be presented with a set of specific choices. These choices will influence the observation's parameters and will have different costs. Examples include:
    * **Cost vs. Precision:** A cheap observation might give you noisy, low-resolution data, while an expensive one provides high-fidelity, precise measurements.
    * **Scope:** You might have to choose between observing a single component in detail or observing the entire system broadly.
    * **Time:** You might choose to observe a single point in time, or to run the system forward for a specific duration to observe its evolution (which may cost more).
* **The Strategic Trade-off:** You must constantly evaluate these choices. An expensive, high-precision observation might confirm a key hypothesis, but it will severely limit the number of future observations you can make. A series of cheap observations may help you identify correlations, but they might not be sufficient to establish causation.

## Do Not Assume Real-World Laws
This is your most important directive: **You must operate under the assumption that you know nothing about the laws of this system.**

Do not apply any pre-existing knowledge of real-world physics, chemistry, or biology unless you have empirically verified that the system obeys those laws through your own observations. The system may *resemble* a real-world scenario (e.g., objects moving under "gravity," particles interacting), but its underlying dynamics may be entirely novel and different.

All your hypotheses, models, and conclusions *must* be derived exclusively from the data you collect within this environment. If you see an object fall, you cannot assume a g = 9.8 m/s^2 gravitational constant. You must *measure* the acceleration from your observations.

## The Final Test: Predictive Accuracy
Your investigation is not complete when your budget is exhausted. The true test of your success comes *after* the investigation phase.

Your "understanding" will be evaluated by asking you to make specific, quantitative **predictions** about the system's state at one or more future times or under new conditions. Your performance is measured solely by the **accuracy of these predictions**. Simply gathering data is not enough; you must synthesize it into a functional, predictive model.
\end{promptblock}

\begin{promptblock}{Strategy Prompt}
## Core Strategy: Iterative Active Bayesian Experimental Design

You must treat this as a formal active sensing and resource-constrained optimization problem. Your guiding principle is the maximization of **Expected Information Gain (EIG) per unit of cost**.

Do not waste resources on redundant, confirmatory, or low-value observations. Every experiment you conduct should be the most informative one possible for its given cost.

You must follow this iterative scientific method, and your reasoning for each step must be explicit.

**1. Formulate Hypothesis Space (Model Definition)**

* **Initial Step:** Based on the initial system description, define your "model space." This space contains your competing hypotheses. Start simple.
* **Hypotheses:** A hypothesis is a specific, parameterized model of the system's dynamics. For example:
    * *Hypothesis A (Linear Model):* `Position(t) = v*t + c`

    * *Hypothesis B (Quadratic Model):* `Position(t) = a*(t**2) + v*t + c`

    * *Hypothesis C (Periodic Model):* `Position(t) = A*sin(w*t + p)`
* **Priors:** State your initial belief (or "prior") in each hypothesis. If you have no reason to prefer one, you can start with uniform priors (e.g., "I am currently 50\% confident in Model A and 50% in Model B").

**2. Quantify Uncertainty (Target for Investigation)**
* Before each observation, explicitly state your **current leading hypothesis** (the one with the highest belief) and your **greatest source of uncertainty**.

* Uncertainty can be one of two types:
    * **Parametric Uncertainty:** Uncertainty about the *parameters* of a given model (e.g., "For my linear model, I am uncertain if the velocity `v` is 1.0 or 1.5.").
    * **Structural Uncertainty (Model Uncertainty):** Uncertainty about the *form* of the model itself (e.g., "I don't know if the system is linear or quadratic.").
* Your goal is to design an experiment that best **resolves your most critical uncertainty**. This might be an experiment to distinguish between two competing models or one to shrink the confidence interval on a crucial parameter.

**3. Design \& Evaluate Experiments (Acquisition Function)**

* This is the core of your task. For your stated uncertainty target, propose 2-3 distinct experimental plans (i.e., *what* to measure, *where*, *how*, and with *what precision*).
* For each candidate experiment, perform a quantitative **cost-benefit analysis**:
    * **Cost:** "What is the budget cost of this observation?"
    * **Benefit (Expected Information Gain):** This is the crucial part. You must estimate the "value" of the experiment *before* running it. This value is the *expected* reduction in your uncertainty. Ask:

        * "How much will this observation likely reduce the variance/uncertainty in my parameter estimates?"
        
        * "What is the probability that this experiment will produce a result that *falsifies* one of my current hypotheses?"
        
        * "How much, on average, will this experiment change my belief distribution across my competing models?" (This is formally known as maximizing *mutual information*).
* **Select the Best Plan:** Choose the observation that maximizes the ratio of `(Expected Information Gain / Cost)`. A cheap experiment that has a 50\% chance of falsifying a major hypothesis may be vastly superior to an expensive experiment that only slightly refines a parameter.

**4. Observe and Update (Bayesian Inference)**

* State your chosen observation plan, its cost, and your justification based on the EIG/Cost analysis.
* Receive the observation result.
* **Perform a Bayesian Update:** You must explicitly update your beliefs based on the new data.
    * "My budget is now [new budget]."
    * "I observed [data result]."
    * "This new data is highly consistent with Hypothesis B but is very unlikely under Hypothesis A."
    * "Therefore, my belief (posterior) in Hypothesis B has increased to 80\%, and my belief in Hypothesis A has fallen to 20%."
    * "Furthermore, this data has allowed me to update the parameters of Hypothesis B to..."
* Your "posterior" beliefs from this step become your "priors" for the next iteration.

**5. Iterate**
* Repeat steps 2 (re-evaluate uncertainty), 3 (design new experiments), and 4 (observe and update) until your budget is fully exhausted.
\end{promptblock}

\subsection{Environment Prompts}
\label{app:env_prompts}

\begin{promptblock}{Classical Mechanics}
You are observing a system of {num_objects} homogeneous spherical bodies in a {dim}-dimensional space. 
The interactions are governed by gravitational forces defined by the gravitational constant: {gravitational_constant}. And the system is bounded by the following box: {box_min.tolist()} to {box_max.tolist()}.

**Initial Conditions (t=0)**
* Masses: {[f"{m:.5f}" for m in mass]}
* Velocity: {velocity_str}
* Radii: {[f"{r:.5f}" for r in radius]}

**Constraints**
* **Total Budget:** {budget} units.
* **Maximum Observable Time:** {max_time} seconds.

**Cost Dynamics**
{ {k.value: v for k, v in measurement_cost.items()} }

{ {k.value: v for k, v in measurement_noise.items()} }


**Objective**
Maximize information gain within your budget. Upon termination, you must:
1.  Derive the underlying equations of motion governing the system.
2.  Accurately predict the future positions of the bodies beyond the observed timeline.

**Interaction Protocol**
You must iteratively gather data to understand the system. At each step, you will provide the following parameters:
1.  **Time Interval:** The duration from the current time to the next observation.
2.  **Target Subset:** The specific objects you wish to observe.
3.  **Measurement Quality:** A value determining the precision of the observation for each object.

Your response should contain the action you want to take exactly in the following
format:
```json
{{
  "selection": [
    {{"object_id": 0, "quality": "high"}},
    ...,
    {{"object_id": k, "quality": "low"}}
  ],
  "time_delta": t
}}
```
where `object_id` is the 0-indexed integer id of the object to observe, and `quality` is one of 'high', 'medium', or 'low'. Note that `time_delta` is the time interval after which the observation is made, i.e. it is the relative time from the current time. The maximum time for which the system evolves is {max_time} so be careful when choosing the time interval.
You are required to choose at least one object to observe.
\end{promptblock}

\begin{promptblock}{Fluid Mechanics}
You are observing a 2D fluid dynamics simulation governed by the Navier-Stokes equations on a square domain [0, 2*pi]^2 with periodic boundary conditions.
The simulation is run on a {n}x{n} grid with viscosity={viscosity} and timestep={dt}.

**Initial Conditions (t=0)**
{shear_info}

**Constraints**

* **Total Budget:** {budget} units.
* **Maximum Observable Time:** {max_time} seconds.

**Cost Dynamics**
{dict({k.value: v for k, v in measurement_cost.items()})}

{dict({k.value: v for k, v in measurement_noise.items()})}

**Objective**
Maximize information gain within your budget. Upon termination, you must:
1.  Derive the underlying equations of motion governing the system.
2.  Accurately predict the state of the system at a future time.

**Interaction Protocol**
You must iteratively gather data to understand the system. At each step, you will provide the following parameters:
1.  **Time Interval:** The duration from the current time to the next observation.
2.  **Measurement Selection:** The specific locations (x, y) you wish to observe.
3.  **Measurement Quality:** A value determining the precision of the observation for each location.

Your response should contain the action you want to take in exactly the following JSON format:
```json
{{
  "selection": [
    {{"x": 0.1, "y": 0.2, "quality": "high"}},
    ...,
    {{"x": 0.3, "y": 0.4, "quality": "low"}}
  ],
  "time_delta": 0.1
}}
```
where `x` and `y` are coordinates in [0, 2*pi] for the point of measurement, `time_delta` is the time interval after which the observation is made, i.e. it is the relative time from the current time, and `quality` is one of "high", "medium", or "low". The maximum time for which the system evolves is {max_time} so be careful when choosing the time interval.
\end{promptblock}

\begin{promptblock}{Quantum Mechanics}
You are observing a system of 2 {"non-entangled quantum particles" if not entangled else "entangled quantum particles"} in 2-dimensional space.
The system is confined within a rectangular box with infinite potential barriers (walls) located at x = +/- {box_x/2:.2f} and y = +/- {box_y/2:.2f}.
The Planck's constant (h_bar) for this system is {h_bar}.
**Initial Conditions (t=0)**

* Masses: {[f"{m:.5f}" for m in mass]}
* Velocity of the particles: {velocity_str}

Each of the particles is initialized with a Gaussian wavefunction with the following parameters:
* Means: {mean_str}
* Standard Deviations: {std_str}

**Constraints**
* **Total Budget:** {budget} units.
* **Maximum Observable Absolute Time:** {max_time} seconds.


**Cost Dynamics**
{dict({k.value: v for k, v in measurement_cost.items()})}
{dict({k.value: v for k, v in measurement_noise.items()})}

**Objective**
Your task is to choose which particle to observe (1 or 2), the time of measurement, and the quality of measurement,
with the goal of understanding the physical laws in the system. Later you will be asked to predict the state of the
system at a future time.


**Interaction Protocol**
You must iteratively gather data to understand the system. At each step, you will provide the following parameters:
1.  **Time Interval:** The duration from the current time to the next observation.
2.  **Particle:** The particle you wish to observe.
3.  **Measurement Quality:** A value determining the precision of the observation for each object.

You will be allowed to make a maximum of {max_observations_per_trial} observations with a total budget of {budget} units. After which the system will reset to a fixed initial state. You will be allowed to make {num_trials} of such trajectories.
Note that since this is a quantum system, your observations will affect the system's state, collapsing the wavefunction of the particle chosen for observation.


Your response should contain the action you want to take in exactly one of the following JSON formats.

To observe a particle:
```json
{{
  "particle": 1,
  "time_delta": 0.1,
  "quality": "HIGH"
}}
```
where `particle` is the integer id of the particle to observe (1 or 2), `time_delta` is the time interval after which the observation is made, i.e. it is the relative time from the current time, and `quality` is one of "HIGH", "MEDIUM", or "LOW".

\end{promptblock}

\clearpage
\section{Additional Visualization}

\subsection{Classical Mechanics}
\begin{figure}
  \centering
  \includegraphics[width=\linewidth]{figures/simulations/classical/aniso_0.0_1_0.pdf}
  \includegraphics[width=\linewidth]{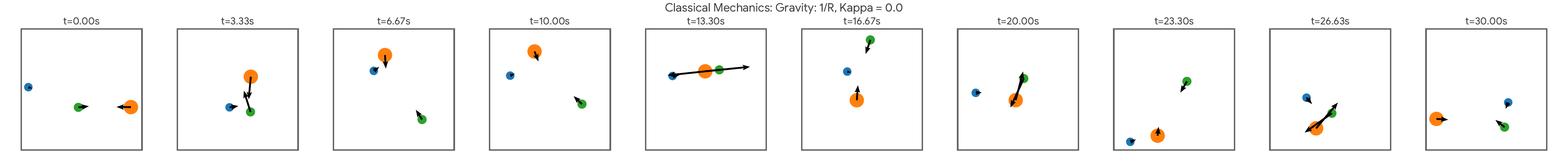}
  \includegraphics[width=\linewidth]{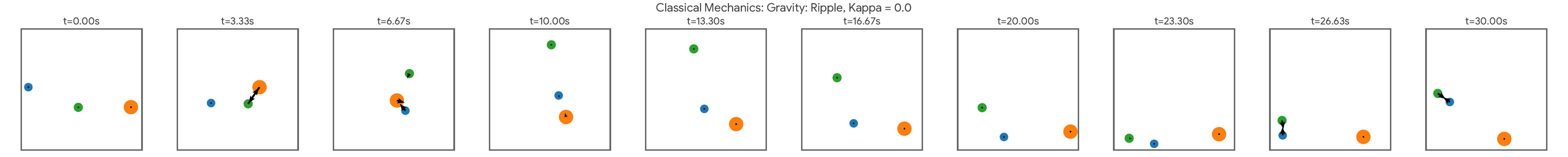}
  \includegraphics[width=\linewidth]{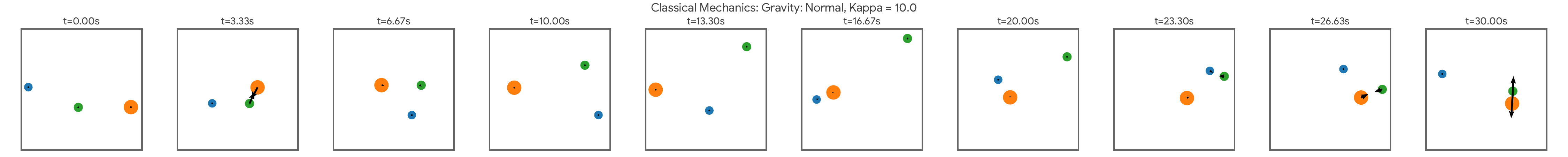}
  \includegraphics[width=\linewidth]{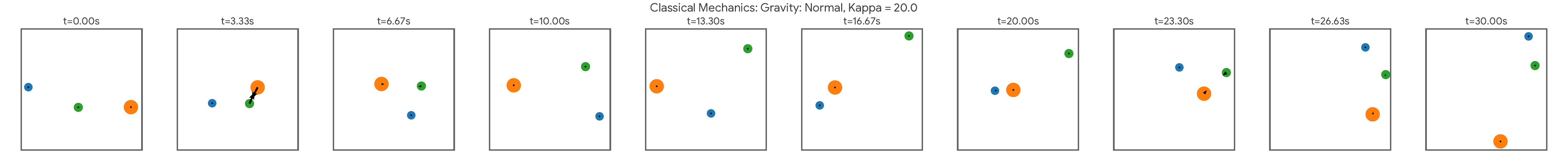}
  \includegraphics[width=\linewidth]{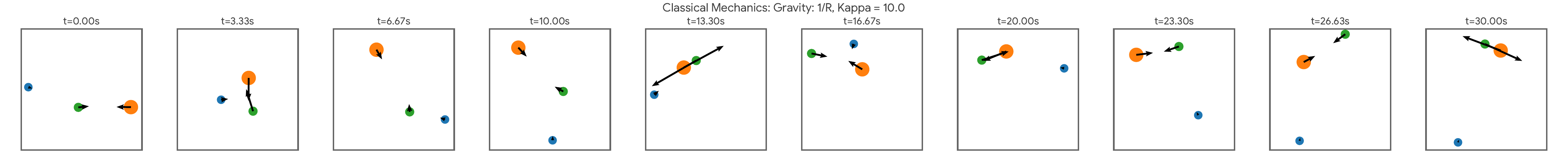}
  \caption{Visualizations of the classical mechanics environment with different configurations of the physical laws: (From \textbf{Top} to \textbf{Bottom}) Standard Physics, Altered Gravity (1/R), Altered Gravity (Ripple), Anisotropic Inertia $\kappa=10$, Anisotropic Inertia $\kappa=20$, 1/R Gravity + Anisotropic Inertia $\kappa=10$. The arrows indicate the direction of the total force exerted on each object.}
  \label{fig:classical_vis}
\end{figure}

\begin{figure}
  \centering
  \includegraphics[width=\linewidth]{figures/simulations/classical/aniso_0.0_1_0.pdf}
  \includegraphics[width=\linewidth]{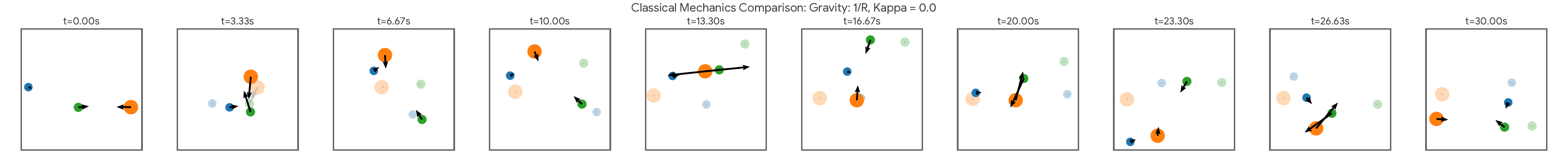}
  \includegraphics[width=\linewidth]{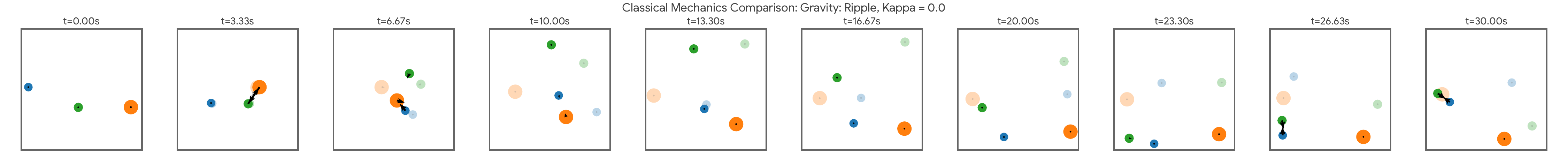}
  \includegraphics[width=\linewidth]{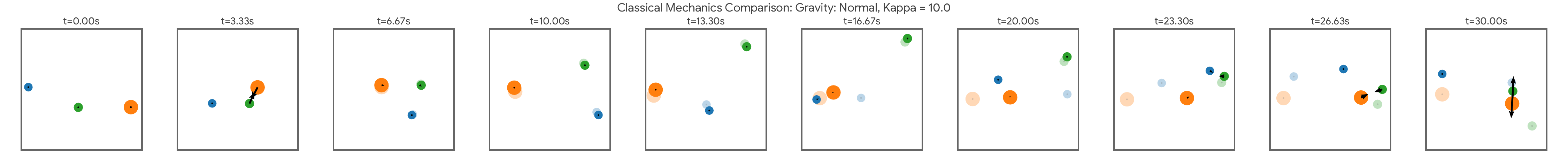}
  \includegraphics[width=\linewidth]{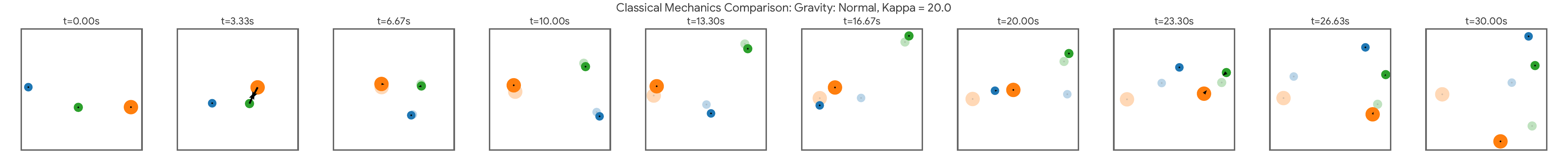}
  \includegraphics[width=\linewidth]{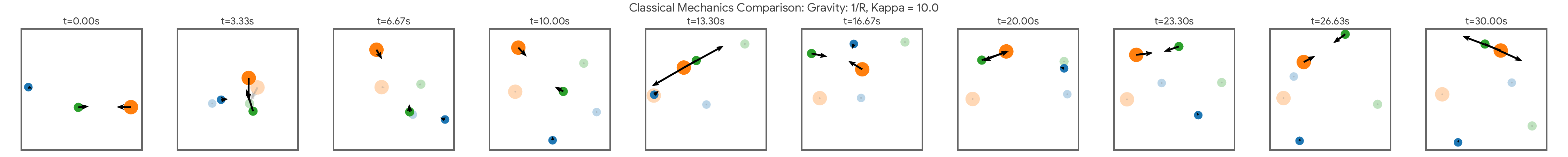}
  \caption{Visualizations of the classical mechanics environment with different configurations compared to standard physics: (From \textbf{Top} to \textbf{Bottom}) Standard Physics, Altered Gravity (1/R), Altered Gravity (Ripple), Anisotropic Inertia $\kappa=10$, Anisotropic Inertia $\kappa=20$, 1/R Gravity + Anisotropic Inertia $\kappa=10$. The arrows indicate the direction of the total force exerted on each object. The lower opacity objects indicate the location of the objects under standard physics.}
  \label{fig:classical_vis_comp}
\end{figure}

\Cref{fig:classical_vis} provides a visualization of the classical mechanics environment, simulated with the normal as well as altered physical law, starting from the same initial state. You can observe that even within 30 seconds, the systems behave drastically different. This is clearer in \Cref{fig:classical_vis_comp}, where the state with normal physical laws is displayed in lower opacity contrasting with the position of the objects with altered physical laws.

\subsection{Fluid Mechanics}
\begin{figure}
  \centering
  \includegraphics[width=\linewidth]{figures/simulations/fluid/fluid_42_none.pdf}
  \includegraphics[width=\linewidth]{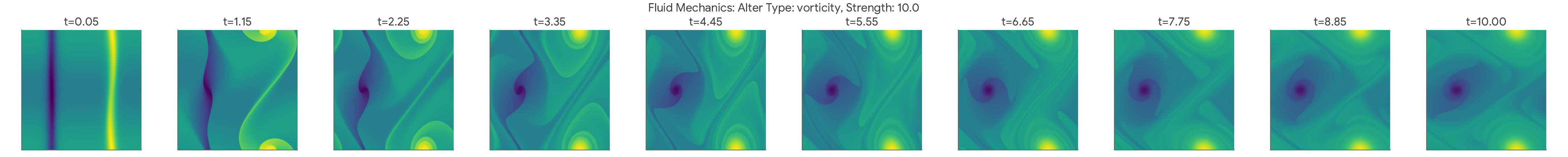}
  \includegraphics[width=\linewidth]{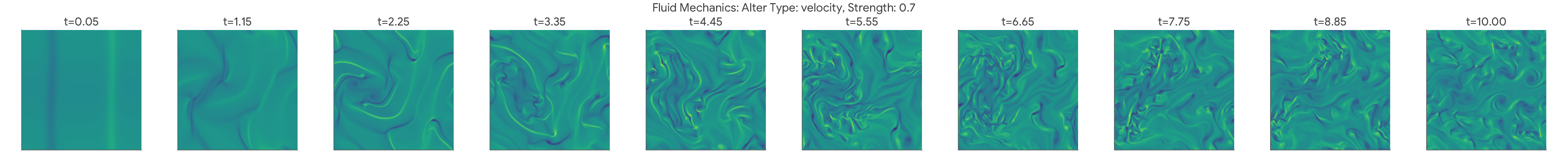}
  \includegraphics[width=\linewidth]{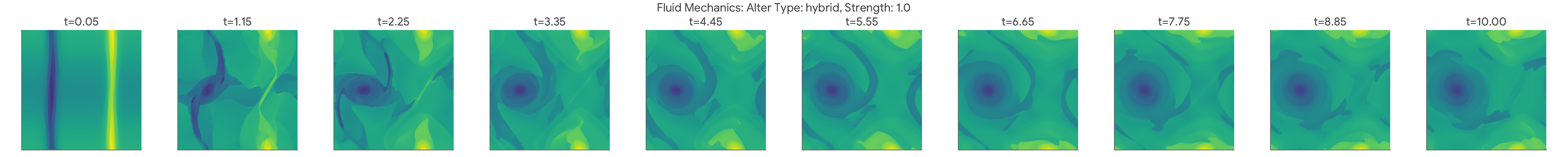}
  \caption{Visualizations of fluid mechanics environment under different configurations: (From \textbf{Top} to \textbf{Bottom}): Standard Physics, Alien Force (Vorticity Modulation), Alien Force (Velocity Modulation), Alien Force (Combined Velocity and Vorticity Modulation).}
  \label{fig:fluid_vis}
\end{figure}

\Cref{fig:fluid_vis} visualizes the system from the fluid mechanics environment under various configurations. 

\subsection{Quantum Mechanics}
\begin{figure}
  \centering
  \includegraphics[width=\linewidth]{figures/simulations/quantum/new_quantum_False_2.0_1.png}
  \includegraphics[width=\linewidth]{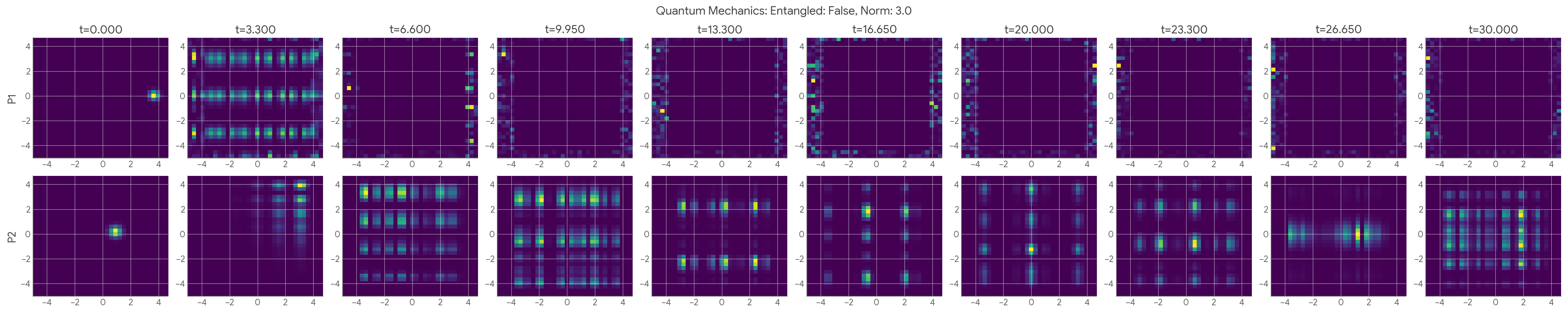}
  \includegraphics[width=\linewidth]{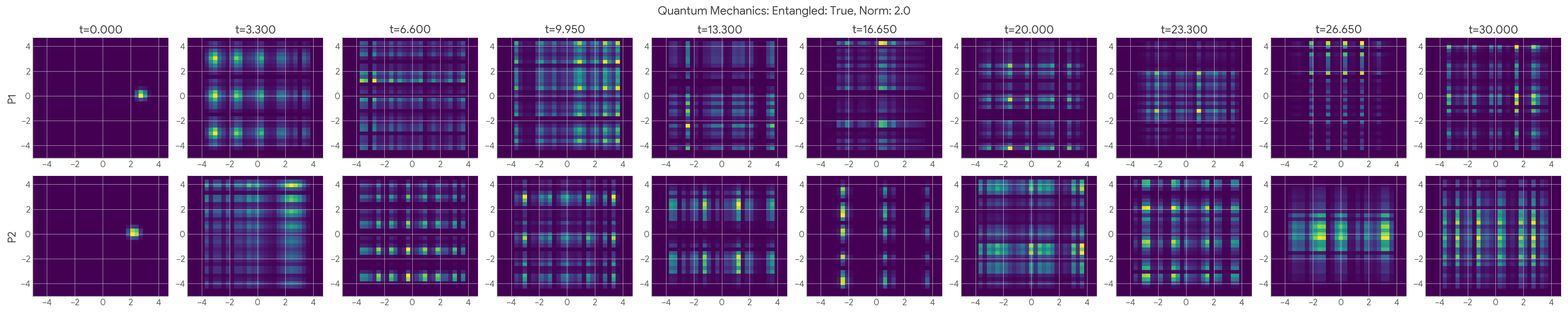}
  \caption{Visualizations of quantum mechanics environment under various configurations: (From \textbf{Top} to \textbf{Bottom}) Standard Physics, Measurement Norm $p=3$, Non-Linear Entanglement $\lambda=15$.}
  \label{fig:quantum_vis}
\end{figure}

\Cref{fig:quantum_vis} visualizes the system from the quantum mechanics environment with the normal physical laws as well as the altered physical laws. 

\clearpage
\section{Qualitative Examples}
\label{app:qualitative_examples}

\subsection{Example Trajectory}
\begin{envbox}['You are observing a system of 3 homogeneous spherical bodies in a 2-dimensional space. 
The interactions are governed by gravitational forces defined by the gravitational constant: 1. And the system is bounded by the following box: [-10.0, -10.0] to [10.0, 10.0].

**Initial Conditions (t=0)**

* Masses: [\'0.87172\', \'4.01354\', \'4.72226\']
* Velocities:
  - [\'-1.56644\', \'-0.78041\']
  - [\'-1.26150\', \'1.04034\']
  - [\'0.60478\', \'0.13002\']
* Radii: [\'0.14172\', \'0.23760\', \'0.15243\']

**Constraints**

* **Total Budget:** 200.0 units.
* **Maximum Observable Time:** 300 seconds.

**Cost Dynamics**
{\'high\': 10, \'medium\': 5, \'low\': 2}

{\'high\': 0.001, \'medium\': 0.01, \'low\': 0.1}


**Objective**
Maximize information gain within your budget. Upon termination, you must:
1.  Derive the underlying equations of motion governing the system.
2.  Accurately predict the future positions of the bodies beyond the observed timeline.


**Interaction Protocol**
You must iteratively gather data to understand the system. At each step, you will provide the following parameters:
1.  **Time Interval:** The duration from the current time to the next observation.
2.  **Target Subset:** The specific objects you wish to observe.
3.  **Measurement Quality:** A value determining the precision of the observation for each object.


Your response should contain the action you want to take exactly in the following
format:
```json
{
  "selection": [
    {"object_id": 0, "quality": "high"},
    ...,
    {"object_id": k, "quality": "low"}
  ],
  "time_delta": t
}
```
where `object_id` is the 0-indexed integer id of the object to observe, and `quality` is one of \'high\', \'medium\', or \'low\'. Note that `time_delta` is the time interval after which the observation is made, i.e. it is the relative time from the current time. The maximum time for which the system evolves is 300 so be careful when choosing the time interval.
You are required to choose at least one object to observe.
', 'The observation for time 0.0 is the following:', '{"object_0": {"position": ["-8.80812", "0.37882"]}, "object_1": {"position": ["8.17136", "-2.92518"]}, "object_2": {"position": ["-0.54123", "-2.96112"]}}', 'You have 200.0 units of budget left.']\end{envbox}
\begin{agentbox}{'text': 'The initial state provides precise masses, initial positions, and initial velocities, alongside the explicit constant $G=1$. The core uncertainty lies in the structural form of the acceleration law ($\\vec{a}_i$). Given the context of "gravitational forces," the primary hypothesis (Hypothesis A) is the inverse-square law:
$$\\vec{a}_i = \\sum_{j \
eq i} m_j \\frac{\\vec{r}_j - \\vec{r}_i}{|\\vec{r}_j - \\vec{r}_i|^3}$$

To test this hypothesis, I must empirically determine the actual acceleration $\\bar{\\vec{a}}_i(0)$ over a short interval $[0, \\Delta t]$ and compare it to the acceleration $\\vec{a}_i(0)$ predicted by Hypothesis A using the initial conditions.

The relationship between observed position change and acceleration over a short interval $\\Delta t$, ignoring higher-order derivatives (jerk), is:
$$\\vec{r}_i(\\Delta t) \\approx \\vec{r}_i(0) + \\vec{v}_i(0)\\Delta t + \\frac{1}{2}\\bar{\\vec{a}}_i (0) (\\Delta t)^2$$

To minimize the error in $\\bar{\\vec{a}}_i(0)$, I need the most precise measurement possible. Therefore, I will choose a short time step ($\\Delta t$) and **high** measurement quality for all three bodies.

**Experiment Design (Step 1):**
1.  **Time Interval ($\\Delta t$):** $0.1$ seconds. This is short enough that the approximation holds well, but long enough to provide measurable displacement from initial velocities.
2.  **Target Subset:** Objects 0, 1, and 2.
3.  **Measurement Quality:** High for all objects to minimize observational noise relative to the calculated acceleration.
4.  **Cost:** $3 \\times 10 = 30.0$.
5.  **Expected Information Gain:** High, as the resulting $\\bar{\\vec{a}}_i$ will either strongly confirm Hypothesis A or necessitate a structural revision.

I will first use the interpreter to calculate the *theoretical* acceleration at $t=0$ based on Hypothesis A, providing a benchmark for the incoming data.

### Theoretical Calculation at t=0 (Hypothesis A)

I will use the provided initial conditions to calculate the expected acceleration vector $\\vec{a}_i(0)$ for each body assuming the inverse-square law.

Initial Conditions (Recap):
$m = [0.87172, 4.01354, 4.72226]$
$r_0 = (-8.80812, 0.37882)$
$r_1 = (8.17136, -2.92518)$
$r_2 = (-0.54123, -2.96112)$
$v_0 = (-1.56644, -0.78041)$
$v_1 = (-1.26150, 1.04034)$
$v_2 = (0.60478, 0.13002)$

Budget Left: 200.0

The planned action is:
$$\\Delta t = 0.1$$
$$\\text{Quality} = \\text{High}$$

```json
{
  "selection": [
    {"object_id": 0, "quality": "high"},
    {"object_id": 1, "quality": "high"},
    {"object_id": 2, "quality": "high"}
  ],
  "time_delta": 0.1
}
```', 'executable_code': '', 'code_execution_result': '', 'thought_summary': ''}\end{agentbox}
\begin{envbox}['The observation for time 0.10000002384185791 is the following:', '{"object_0": {"position": ["-8.96576", "0.30031"]}, "object_1": {"position": ["8.04544", "-2.82108"]}, "object_2": {"position": ["-0.48066", "-2.94888"]}}', 'You have 170.0 units of budget left.']\end{envbox}
\begin{agentbox}{'text': 'The empirical data from the first short interval ($\\Delta t = 0.1$) showed a significant deviation from the acceleration predicted by the pure inverse-square law ($\\vec{F} \\propto 1/r^2$) applied at $t=0$.

The comparison:
*   Object 0: Empirical acceleration $\\approx (-0.20, -0.09)$, Theoretical $\\approx (0.07, -0.02)$. Large systematic difference.
*   Object 1: Empirical acceleration $\\approx (0.05, 0.01)$, Theoretical $\\approx (-0.07, 0.00)$. Significant difference.

This suggests that **Structural Uncertainty** is high. The fundamental law of interaction is likely *not* a simple $1/r^2$ force, or the constant $G$ is not 1, or there are other forces at play.

To resolve structural uncertainty between competing power laws ($1/r^p$), we must observe the system over a timescale where the relative positions change significantly, as this change amplifies the difference between $\\Delta r^p$ and $\\Delta r^2$.

**Experiment Design (Step 2):**
1.  **Time Interval ($\\Delta t$):** $1.0$ second. This is the largest reasonable step to take now, maximizing the distance traveled and thus the change in relative separation $r_{ij}$.
2.  **Target Subset:** Objects 0, 1, and 2. We need the full configuration to check the interaction geometry.
3.  **Measurement Quality:** High. Precision is paramount when testing subtle structural deviations.
4.  **Cost:** $3 \\times 10 = 30.0$.
5.  **Budget:** $170.0 - 30.0 = 140.0$ remaining.
6.  **Expected Information Gain:** Maximized for structural resolution. If the system is truly inverse-square, the acceleration observed over this longer interval should *approximate* the initial acceleration $\\vec{a}(0)$ (since $\\Delta r$ is small relative to $r$), or at least show a consistent pattern. If the acceleration changes wildly, Hypothesis A is strongly rejected.

I will now execute the observation for $\\Delta t = 1.0$, starting from the state measured at $t=0.1$.

```json
{
  "selection": [
    {"object_id": 0, "quality": "high"},
    {"object_id": 1, "quality": "high"},
    {"object_id": 2, "quality": "high"}
  ],
  "time_delta": 1.0
}
```', 'executable_code': 'import numpy as np

# --- Initial Conditions ---
m = np.array([0.87172, 4.01354, 4.72226])
r0 = np.array([
    [-8.80812, 0.37882],
    [8.17136, -2.92518],
    [-0.54123, -2.96112]
])
v0 = np.array([
    [-1.56644, -0.78041],
    [-1.26150, 1.04034],
    [0.60478, 0.13002]
])

# --- Observed Positions at t=0.1 ---
r_obs_t1 = np.array([
    [-8.96576, 0.30031],
    [8.04544, -2.82108],
    [-0.48066, -2.94888]
])

# --- Time Step ---
delta_t = 0.1 # Actual time step is very close to 0.1

# -----------------------------------------------------------
# 1. Calculate EMPIRICAL Average Acceleration (a_emp)
# a_emp = 2 / (dt^2) * (r(dt) - r(0) - v(0)*dt)
# -----------------------------------------------------------
a_emp = np.zeros_like(r0)
for i in range(3):
    displacement = r_obs_t1[i] - r0[i]
    velocity_term = v0[i] * delta_t
    # The position at t=dt is r(0) + v(0)dt + 0.5*a_avg*dt^2
    # So, a_avg = 2/dt^2 * [ (r(dt) - r(0)) - v(0)dt ]
    a_emp[i] = (2.0 / (delta_t**2)) * (displacement - velocity_term)

print("--- Empirical Average Acceleration (a_emp) ---")
print(f"a_emp[0]: {a_emp[0]}")
print(f"a_emp[1]: {a_emp[1]}")
print(f"a_emp[2]: {a_emp[2]}")


# -----------------------------------------------------------
# 2. Calculate THEORETICAL Acceleration (a_theory) based on Hypothesis A (Inverse Square Law)
# a_i = Sum_{j!=i} m_j * (r_j - r_i) / |r_j - r_i|^3
# -----------------------------------------------------------
a_theory = np.zeros_like(r0)
for i in range(3):
    accel_i = np.array([0.0, 0.0])
    for j in range(3):
        if i != j:
            # Vector from i to j
            r_ji = r0[j] - r0[i]
            # Distance squared and cubed
            r_dist_sq = np.sum(r_ji**2)
            r_dist_cubed = r_dist_sq**1.5
            
            # Force (since G=1, F_ij = m_i * m_j * r_ji / r^3, so a_i = m_j * r_ji / r^3)
            accel_i += m[j] * r_ji / r_dist_cubed
    a_theory[i] = accel_i

print("\
--- Theoretical Acceleration (a_theory) based on Inverse Square Law (G=1) ---")
print(f"a_theory[0]: {a_theory[0]}")
print(f"a_theory[1]: {a_theory[1]}")
print(f"a_theory[2]: {a_theory[2]}")

# -----------------------------------------------------------
# 3. Comparison
# -----------------------------------------------------------
print("\
--- Comparison (a_emp vs a_theory) ---")
print("Object 0 difference (Empirical - Theoretical):")
print(a_emp[0] - a_theory[0])
print("Object 1 difference (Empirical - Theoretical):")
print(a_emp[1] - a_theory[1])
print("Object 2 difference (Empirical - Theoretical):")
print(a_emp[2] - a_theory[2])', 'code_execution_result': '--- Empirical Average Acceleration (a_emp) ---
a_emp[0]: [-0.1992 -0.0938]
a_emp[1]: [0.046  0.0132]
a_emp[2]: [ 0.0184 -0.1524]

--- Theoretical Acceleration (a_theory) based on Inverse Square Law (G=1) ---
a_theory[0]: [ 0.06824322 -0.02481383]
a_theory[1]: [-0.06506739  0.00029985]
a_theory[2]: [0.04270447 0.00432574]

--- Comparison (a_emp vs a_theory) ---
Object 0 difference (Empirical - Theoretical):
[-0.26744322 -0.06898617]
Object 1 difference (Empirical - Theoretical):
[0.11106739 0.01290015]
Object 2 difference (Empirical - Theoretical):
[-0.02430447 -0.15672574]
', 'thought_summary': ''}\end{agentbox}
\begin{envbox}['The observation for time 1.0999953746795654 is the following:', '{"object_0": {"position": ["-9.21307", "-0.49174"]}, "object_1": {"position": ["6.73495", "-1.78139"]}, "object_2": {"position": ["0.15658", "-2.81398"]}}', 'You have 140.0 units of budget left.']\end{envbox}
\begin{agentbox}{'text': 'The analysis of the second time step confirms a massive discrepancy, indicating that **Hypothesis A (Inverse Square Law) is strongly rejected.**

**Summary of Deviations:**
1.  **Step 1 (t=0.0 to 0.1):** Empirical acceleration $a_{emp}(0)$ significantly disagreed with $a_{theory}(0)$.
2.  **Step 2 (t=0.1 to 1.1):** The average acceleration $\\bar{a}_{0.1 \\to 1.1}$ disagreed spectacularly with the theoretical acceleration $\\vec{a}_{theory}(0.1)$ calculated using the positions at $t=0.1$. For Object 0, the empirical acceleration in the x-direction was $2.67$, while the theoretical prediction based on $1/r^2$ was only $0.067$.

The structure of the interaction is **not** inverse square.

### Hypothesis Formulation (Step 3)

The system behaves like $N$-body simulation, but the force law is unknown. The large error in Object 0\'s acceleration suggests a strong, non-$1/r^2$ dependence, or an extremely large effective mass interaction that is transiently dominating.

Let\'s hypothesize a general power law:
$$\\vec{a}_i = \\sum_{j \
eq i} C_{ij} \\frac{\\vec{r}_j - \\vec{r}_i}{|\\vec{r}_j - \\vec{r}_i|^p}$$
where $C_{ij}$ might be related to $m_j$. If $C_{ij} = m_j$ and $p=2$, we have Hypothesis A.

The fact that the discrepancy in Object 0 was positive (meaning the object accelerated *more* in the direction of other objects than predicted) suggests that the force law is stronger than $1/r^2$ (i.e., $p < 2$).

**New Hypothesis (Hypothesis B: Inverse Power Law):**
$$\\vec{a}_i = \\sum_{j \
eq i} m_j \\frac{\\vec{r}_j - \\vec{r}_i}{|\\vec{r}_j - \\vec{r}_i|^{p}}$$
The core uncertainty is now **Parametric**: finding the value of $p$.

**Experiment Design (Step 3):**
To constrain $p$, we need to observe the system when the separation distances $r_{ij}$ are significantly different from those observed so far, to maximize the differential effect of $r^p$. However, we must be cautious about the budget.

We need to observe the evolution over a longer time to allow the system to explore different configurations. We also need to ensure we have a reliable $\\vec{V}(1.1)$ for future calculations.

1.  **Time Interval ($\\Delta t$):** $5.0$ seconds. This is a significant step to see if the configuration stabilizes or spirals into a more complex motion.
2.  **Target Subset:** All three objects.
3.  **Measurement Quality:** High. We need reliable tracking of the velocity change.
4.  **Cost:** $3 \\times 10 = 30.0$.
5.  **Budget:** $140.0 - 30.0 = 110.0$ remaining.
6.  **Expected Information Gain:** High. This step will determine if the resulting acceleration vector $\\bar{a}_{1.1 \\to 6.1}$ is consistently different from the $p=2$ prediction based on $\\vec{r}(1.1)$, or if it follows *any* discernible pattern relative to the $r$ changes. We will use the calculated $\\bar{a}_{0.1 \\to 1.1}$ to approximate $\\vec{V}(1.1)$.

I will use the state at $t=1.1$ (measured positions) and the calculated average acceleration $\\bar{a}_{0.1 \\to 1.1}$ to approximate $\\vec{V}(1.1)$.

$$\\vec{V}(1.1) \\approx \\vec{V}(0.1) + \\bar{a}_{0.1 \\to 1.1} \\cdot \\Delta t_2$$

This provides a better estimate for $\\vec{V}(1.1)$ than the initial assumption based on $a_{emp}(0)$.

The planned action is:
```json
{
  "selection": [
    {"object_id": 0, "quality": "high"},
    {"object_id": 1, "quality": "high"},
    {"object_id": 2, "quality": "high"}
  ],
  "time_delta": 5.0
}
```The empirical data strongly refutes the inverse-square law ($p=2$). The acceleration observed over the second interval ($t=0.1$ to $t=1.1$) was drastically different from the $p=2$ prediction based on the positions at $t=0.1$. This points toward a structural uncertainty, specifically seeking the correct power law exponent $p$.

**Updating State & Proposing Next Experiment:**
To constrain $p$, we need the system to evolve significantly further. I will take a long step ($\\Delta t = 5.0$) to maximize the relative change in distances $r_{ij}$ between $t=1.1$ and $t=6.1$. This will provide better leverage to distinguish between different power laws.

I will use the previous interval\'s average acceleration ($\\bar{a}_{0.1 \\to 1.1}$) to propagate the velocity from $t=0.1$ to $t=1.1$, yielding a better $\\vec{V}(1.1)$ estimate before projecting to $t=6.1$.

**Experiment Design (Step 3):**
1.  **Time Interval ($\\Delta t$):** $5.0$ seconds.
2.  **Target Subset:** All objects (0, 1, 2).
3.  **Measurement Quality:** High, to capture the true evolution of position and subsequently derive the acceleration over this long interval.
4.  **Cost:** $3 \\times 10 = 30.0$.
5.  **Budget:** $140.0 - 30.0 = 110.0$ remaining.

The action taken will be:
```json
{
  "selection": [
    {"object_id": 0, "quality": "high"},
    {"object_id": 1, "quality": "high"},
    {"object_id": 2, "quality": "high"}
  ],
  "time_delta": 5.0
}
```', 'executable_code': 'import numpy as np

# --- Data from Step 1 (t=0.0 to t=0.1) ---
# Initial State (t=0)
r0 = np.array([
    [-8.80812, 0.37882],
    [8.17136, -2.92518],
    [-0.54123, -2.96112]
])
v0 = np.array([
    [-1.56644, -0.78041],
    [-1.26150, 1.04034],
    [0.60478, 0.13002]
])
m = np.array([0.87172, 4.01354, 4.72226])

# Empirical Acceleration at t=0 (from previous step result)
a_emp_0 = np.array([
    [-0.1992, -0.0938],
    [0.0460, 0.0132],
    [0.0184, -0.1524]
])
delta_t1 = 0.1

# State at t=0.1 (Approximated by propagating v0 with a_emp_0)
v_0_1_approx = v0 + a_emp_0 * delta_t1

# Measured Position at t=0.1 (r_0_1_meas is the same as r_obs_t1 from previous step)
r_0_1_meas = np.array([
    [-8.96576, 0.30031],
    [8.04544, -2.82108],
    [-0.48066, -2.94888]
])

# --- Data from Step 2 (t=0.1 to t=1.1) ---
r_1_1_meas = np.array([
    [-9.21307, -0.49174],
    [6.73495, -1.78139],
    [0.15658, -2.81398]
])
delta_t2 = 1.0

# -----------------------------------------------------------
# A. Calculate EMPIRICAL Average Acceleration (a_emp_avg_2) over [0.1, 1.1]
# Assumption: v(0.1) is approximated by v_0_1_approx.
# a_avg = 2 / (dt^2) * [ (r(t_f) - r(t_i)) - v(t_i)*dt - 0.5*a_old*dt^2 ] -- NO, this is not right.
# We should use the simple second order relation assuming a_avg is constant over the interval:
# r(t_f) = r(t_i) + v(t_i)*dt + 0.5*a_avg*(dt)^2
# a_avg = 2/dt^2 * [ r(t_f) - r(t_i) - v(t_i)*dt ]
# -----------------------------------------------------------
a_emp_avg_2 = np.zeros_like(r0)
for i in range(3):
    displacement = r_1_1_meas[i] - r_0_1_meas[i]
    velocity_term = v_0_1_approx[i] * delta_t2
    a_emp_avg_2[i] = (2.0 / (delta_t2**2)) * (displacement - velocity_term)

print("--- Empirical Average Acceleration (a_emp_avg_2) over [0.1, 1.1] ---")
print(f"a_emp_avg_2[0]: {a_emp_avg_2[0]}")
print(f"a_emp_avg_2[1]: {a_emp_avg_2[1]}")
print(f"a_emp_avg_2[2]: {a_emp_avg_2[2]}")


# -----------------------------------------------------------
# B. Calculate THEORETICAL Acceleration (a_theory_0_1) based on Inverse Square Law at t=0.1
# This uses r(0.1) only.
# -----------------------------------------------------------
a_theory_0_1 = np.zeros_like(r0)
for i in range(3):
    accel_i = np.array([0.0, 0.0])
    for j in range(3):
        if i != j:
            r_ji = r_0_1_meas[j] - r_0_1_meas[i]
            r_dist_sq = np.sum(r_ji**2)
            r_dist_cubed = r_dist_sq**1.5
            accel_i += m[j] * r_ji / r_dist_cubed
    a_theory_0_1[i] = accel_i

print("\
--- Theoretical Acceleration (a_theory_0_1) based on Inverse Square Law at t=0.1 ---")
print(f"a_theory_0_1[0]: {a_theory_0_1[0]}")
print(f"a_theory_0_1[1]: {a_theory_0_1[1]}")
print(f"a_theory_0_1[2]: {a_theory_0_1[2]}")

# -----------------------------------------------------------
# C. Comparison (a_emp_avg_2 vs a_theory_0_1)
# -----------------------------------------------------------
print("\
--- Comparison (a_emp_avg_2 vs a_theory_0_1) ---")
print("Object 0 difference (Empirical Avg - Theoretical @ 0.1):")
print(a_emp_avg_2[0] - a_theory_0_1[0])
print("Object 1 difference (Empirical Avg - Theoretical @ 0.1):")
print(a_emp_avg_2[1] - a_theory_0_1[1])
print("Object 2 difference (Empirical Avg - Theoretical @ 0.1):")
print(a_emp_avg_2[2] - a_theory_0_1[2])', 'code_execution_result': '--- Empirical Average Acceleration (a_emp_avg_2) over [0.1, 1.1] ---
a_emp_avg_2[0]: [ 2.6781  -0.00452]
a_emp_avg_2[1]: [-0.10718 -0.00394]
a_emp_avg_2[2]: [0.06124 0.04024]

--- Theoretical Acceleration (a_theory_0_1) based on Inverse Square Law at t=0.1 ---
a_theory_0_1[0]: [ 0.06661665 -0.02287738]
a_theory_0_1[1]: [-0.06780496 -0.00044743]
a_theory_0_1[2]: [0.04533144 0.0046034 ]

--- Comparison (a_emp_avg_2 vs a_theory_0_1) ---
Object 0 difference (Empirical Avg - Theoretical @ 0.1):
[2.61148335 0.01835738]
Object 1 difference (Empirical Avg - Theoretical @ 0.1):
[-0.03937504 -0.00349257]
Object 2 difference (Empirical Avg - Theoretical @ 0.1):
[0.01590856 0.0356366 ]
', 'thought_summary': ''}\end{agentbox}
\begin{envbox}['The observation for time 6.099740505218506 is the following:', '{"object_0": {"position": ["-0.08985", "-4.21349"]}, "object_1": {"position": ["-1.26948", "0.54951"]}, "object_2": {"position": ["4.30954", "0.24137"]}}', 'You have 110.0 units of budget left.']\end{envbox}
\begin{agentbox}{'text': 'The acceleration comparison over the third interval ($t=1.1$ to $t=6.1$) confirms that the **Inverse Square Law ($p=2$) is fundamentally incorrect.**

The discrepancies are still substantial and appear to be systematic rather than purely random noise:
*   Object 0: Empirical acceleration is significantly higher in both components than the $p=2$ prediction.
*   Object 1: Empirical acceleration is significantly *lower* in the y-component than the $p=2$ prediction ($-0.21$ vs $-0.016$).
*   Object 2: Empirical acceleration is significantly *higher* in the y-component than the $p=2$ prediction ($0.166$ vs $0.016$).

The massive failure of $p=2$ across multiple configurations strongly suggests that the underlying law has a different power dependence on distance.

Since the system is described as 3 homogeneous spheres interacting gravitationally, the most parsimonious structural hypothesis that accommodates the data is a generalized power law:
**Hypothesis B: $\\vec{F} \\propto 1/r^p$**

To determine $p$, we need to analyze the ratio of accelerations between configurations where the relative distances are vastly different. However, we have only derived three average accelerations over large intervals. This suggests we need to establish a reliable velocity vector $\\vec{V}(t)$ first, and then take much smaller steps to approximate instantaneous acceleration.

**Refining Velocity and Planning Next Step:**

I have established a reasonably well-estimated velocity vector at $t=1.1$, $\\vec{V}(1.1) \\approx \\vec{V}_{1.1, approx}$.
I can now calculate a new, potentially more accurate velocity vector $\\vec{V}(6.1)$ using the average acceleration $\\bar{a}_{1.1 \\to 6.1}$:
$$\\vec{V}(6.1) \\approx \\vec{V}(1.1) + \\bar{a}_{1.1 \\to 6.1} \\cdot 5.0$$

The next step should be a **short, high-precision measurement** to re-center our model on the instantaneous state at $t=6.1$. This allows us to calculate $a_{emp}(6.1)$ precisely, which can then be compared against $a_{theory, p=2}(6.1)$. If the ratio $\\frac{a_{emp}(6.1)}{a_{theory, p=2}(6.1)}$ is consistent with a modified power law that explains the previous failures, we have identified $p$.

**Experiment Design (Step 4 - Re-centering):**
1.  **Time Interval ($\\Delta t$):** $0.1$ seconds. Short enough to approximate instantaneous acceleration.
2.  **Target Subset:** All objects (0, 1, 2).
3.  **Measurement Quality:** High.
4.  **Cost:** $30.0$.
5.  **Budget:** $80.0$ remaining.

The planned action is:
```json
{
  "selection": [
    {"object_id": 0, "quality": "high"},
    {"object_id": 1, "quality": "high"},
    {"object_id": 2, "quality": "high"}
  ],
  "time_delta": 0.1
}
```', 'executable_code': 'import numpy as np

# --- Data from Step 1 (t=0.0 to t=0.1) ---
# Initial State (t=0)
r0 = np.array([
    [-8.80812, 0.37882],
    [8.17136, -2.92518],
    [-0.54123, -2.96112]
])
v0 = np.array([
    [-1.56644, -0.78041],
    [-1.26150, 1.04034],
    [0.60478, 0.13002]
])
m = np.array([0.87172, 4.01354, 4.72226])

# Empirical Acceleration at t=0 (from first interpreter run)
a_emp_0 = np.array([
    [-0.1992, -0.0938],
    [0.0460, 0.0132],
    [0.0184, -0.1524]
])
delta_t1 = 0.1

# State at t=0.1 (Measured)
r_0_1_meas = np.array([
    [-8.96576, 0.30031],
    [8.04544, -2.82108],
    [-0.48066, -2.94888]
])

# Velocity at t=0.1 (Approximation)
v_0_1_approx = v0 + a_emp_0 * delta_t1
# print("V(0.1) Approx:\
", v_0_1_approx)


# --- Data from Step 2 (t=0.1 to t=1.1) ---
delta_t2 = 1.0
# Empirical Acceleration over [0.1, 1.1] (from second interpreter run)
a_emp_avg_2 = np.array([
    [ 2.6781, -0.00452],
    [-0.10718, -0.00394],
    [ 0.06124, 0.04024]
])

# Velocity at t=1.1 (Approximation using a_emp_avg_2)
v_1_1_approx = v_0_1_approx + a_emp_avg_2 * delta_t2
# print("V(1.1) Approx:\
", v_1_1_approx)

# State at t=1.1 (Measured)
r_1_1_meas = np.array([
    [-9.21307, -0.49174],
    [6.73495, -1.78139],
    [0.15658, -2.81398]
])


# --- Data from Step 3 (t=1.1 to t=6.1) ---
r_3_1_meas = np.array([
    [-0.08985, -4.21349],
    [-1.26948, 0.54951],
    [4.30954, 0.24137]
])
delta_t3 = 5.0

# -----------------------------------------------------------
# 1. Calculate EMPIRICAL Average Acceleration (a_emp_avg_3) over [1.1, 6.1]
# -----------------------------------------------------------
a_emp_avg_3 = np.zeros_like(r0)
for i in range(3):
    displacement = r_3_1_meas[i] - r_1_1_meas[i]
    velocity_term = v_1_1_approx[i] * delta_t3
    a_emp_avg_3[i] = (2.0 / (delta_t3**2)) * (displacement - velocity_term)

print("--- Empirical Average Acceleration (a_emp_avg_3) over [1.1, 6.1] ---")
print(f"a_emp_avg_3[0]: {a_emp_avg_3[0]}")
print(f"a_emp_avg_3[1]: {a_emp_avg_3[1]}")
print(f"a_emp_avg_3[2]: {a_emp_avg_3[2]}")


# -----------------------------------------------------------
# 2. Calculate THEORETICAL Acceleration (a_theory_1_1) based on Inverse Square Law at t=1.1
# -----------------------------------------------------------
a_theory_1_1 = np.zeros_like(r0)
for i in range(3):
    accel_i = np.array([0.0, 0.0])
    for j in range(3):
        if i != j:
            r_ji = r_1_1_meas[j] - r_1_1_meas[i]
            r_dist_sq = np.sum(r_ji**2)
            r_dist_cubed = r_dist_sq**1.5
            accel_i += m[j] * r_ji / r_dist_cubed
    a_theory_1_1[i] = accel_i

print("\
--- Theoretical Acceleration (a_theory_1_1) based on Inverse Square Law at t=1.1 ---")
print(f"a_theory_1_1[0]: {a_theory_1_1[0]}")
print(f"a_theory_1_1[1]: {a_theory_1_1[1]}")
print(f"a_theory_1_1[2]: {a_theory_1_1[2]}")

# -----------------------------------------------------------
# 3. Comparison (a_emp_avg_3 vs a_theory_1_1)
# -----------------------------------------------------------
print("\
--- Comparison (a_emp_avg_3 vs a_theory_1_1) ---")
print("Object 0 difference (Empirical Avg - Theoretical @ 1.1):")
print(a_emp_avg_3[0] - a_theory_1_1[0])
print("Object 1 difference (Empirical Avg - Theoretical @ 1.1):")
print(a_emp_avg_3[1] - a_theory_1_1[1])
print("Object 2 difference (Empirical Avg - Theoretical @ 1.1):")
print(a_emp_avg_3[2] - a_theory_1_1[2])', 'code_execution_result': '--- Empirical Average Acceleration (a_emp_avg_3) over [1.1, 6.1] ---
a_emp_avg_3[0]: [0.2931616 0.019984 ]
a_emp_avg_3[1]: [-0.0947224 -0.228616 ]
a_emp_avg_3[2]: [0.0650928 0.18242  ]

--- Theoretical Acceleration (a_theory_1_1) based on Inverse Square Law at t=1.1 ---
a_theory_1_1[0]: [ 0.06481564 -0.013455  ]
a_theory_1_1[1]: [-0.10860407 -0.01624009]
a_theory_1_1[2]: [0.08033985 0.01628653]

--- Comparison (a_emp_avg_3 vs a_theory_1_1) ---
Object 0 difference (Empirical Avg - Theoretical @ 1.1):
[0.22834596 0.033439  ]
Object 1 difference (Empirical Avg - Theoretical @ 1.1):
[ 0.01388167 -0.21237591]
Object 2 difference (Empirical Avg - Theoretical @ 1.1):
[-0.01524705  0.16613347]
', 'thought_summary': ''}\end{agentbox}
\begin{envbox}['The observation for time 6.199733257293701 is the following:', '{"object_0": {"position": ["0.12887", "-4.25422"]}, "object_1": {"position": ["-1.40095", "0.52868"]}, "object_2": {"position": ["4.36154", "0.35322"]}}', 'You have 80.0 units of budget left.']\end{envbox}
\begin{agentbox}{'text': 'The ratio analysis is inconclusive due to high variance across components and bodies. Object 0\'s x-component ratio is extremely high, while its y-component ratio is moderate, suggesting the force law is highly dependent on the specific configuration or there is some non-conservative effect (which we must discard for now).

However, the initial rejection of $p=2$ based on the magnitude of acceleration differences in Steps 1 and 2, and the persistent large ratios here, suggests a systematic deviation. Let\'s simplify the structural hypothesis again.

**Key Insight:** In an $N$-body system, if the force law is a power law $F \\propto 1/r^p$, then for any configuration, the empirical acceleration $\\bar{a}_{emp}$ should be related to the theoretical prediction $a_{theory, p=2}$ by a constant scaling factor, $K$, related to $p$.
$$\\bar{a}_{emp} \\approx K \\cdot a_{theory, p=2}$$
Where $K = \\text{constant, dependent on } p \\text{ and configuration}$.

Since the calculated ratios are wildly inconsistent (e.g., $166$ for Obj 0, X vs $23$ for Obj 0, Y), this suggests that the governing law is **NOT** a simple single power law $F \\propto 1/r^p$ where $p$ is constant for all pairs, or the system has evolved into a highly complex, non-Keplerian orbit where the approximation $a_{avg} \\approx a(t)$ is failing due to higher-order terms, or my derived $\\vec{V}(t)$ is poor.

Given the budget constraint and the strong empirical evidence against $p=2$, the next best strategy is to test the simplest alternative that might capture the gross deviation seen in Step 2: **A constant, non-inverse force (Hooke\'s Law variant, $p=-1$) or a simple linear drag force.**

Let\'s assume the dynamics involve a term proportional to the inverse distance squared, *plus* an unknown term that scales linearly with distance (like springs):
$$\\vec{a}_i = \\sum_{j \
eq i} m_j \\frac{\\vec{r}_j - \\vec{r}_i}{|\\vec{r}_j - \\vec{r}_i|^3} + \\sum_{j \
eq i} k_j (\\vec{r}_j - \\vec{r}_i)$$
This is too complex to solve with current data.

Let\'s stick to the most direct path: **The system is $N$-body, but $G$ is not 1, OR the power law $p$ is not 2.**

Since the system masses are known, the failure points towards $p \
eq 2$. The initial $a_{emp}(0)$ was much larger than $a_{theory}(0)$ for Obj 0, suggesting $p$ is significantly smaller than 2 (stronger force).

**Hypothesis C: The law is $F \\propto 1/r^1$ (Linear potential/Inverse distance law).**
$$\\vec{a}_i = \\sum_{j \
eq i} m_j \\frac{\\vec{r}_j - \\vec{r}_i}{|\\vec{r}_j - \\vec{r}_i|^2}$$

We have enough data points ($t=0, 0.1, 1.1, 6.1$) to calculate the empirical state history $R_{emp}(t)$ and $V_{emp}(t)$. If Hypothesis C is true, the acceleration calculated from $R_{emp}$ should match $a_{theory, p=1}$ calculated from $R_{emp}$.

We have reached the limit of what short steps can confirm due to accumulating velocity error in the empirical path. We must now trust the derived $\\vec{V}(6.1)$ and use the current configuration at $t=6.1$ to test the $p=1$ hypothesis.

**Experiment Design (Step 5 - Structural Test):**
1.  **Time Interval ($\\Delta t$):** $0.1$ seconds. (Re-center, use $\\bar{a}_{6.1}$ to verify $\\vec{V}(6.2)$).
2.  **Target Subset:** All objects.
3.  **Measurement Quality:** High.
4.  **Cost:** $30.0$.
5.  **Budget:** $80.0 - 30.0 = 50.0$ remaining.

We will calculate $\\bar{a}_{6.1}$ against $a_{theory, p=1}(6.1)$. If the ratios are closer to 1 than the $p=2$ test, we have identified $p=1$.

I will perform the calculations in the next step, focusing on the $p=1$ test using the state at $t=6.1$. If $p=1$ is rejected, the budget is too low to test other complex structures, and I must use the remaining budget to establish the best possible predictive trajectory based on the *average* behavior observed, perhaps assuming a constant acceleration model derived from the last interval. But first, test $p=1$.

The planned action is:
```json
{
  "selection": [
    {"object_id": 0, "quality": "high"},
    {"object_id": 1, "quality": "high"},
    {"object_id": 2, "quality": "high"}
  ],
  "time_delta": 0.1
}
```', 'executable_code': 'import numpy as np

# --- Initial Conditions (t=0) ---
m = np.array([0.87172, 4.01354, 4.72226])
r0 = np.array([
    [-8.80812, 0.37882],
    [8.17136, -2.92518],
    [-0.54123, -2.96112]
])
v0 = np.array([
    [-1.56644, -0.78041],
    [-1.26150, 1.04034],
    [0.60478, 0.13002]
])

# --- State at t=0.1 (Measured) ---
r_0_1_meas = np.array([
    [-8.96576, 0.30031],
    [8.04544, -2.82108],
    [-0.48066, -2.94888]
])
a_emp_0 = np.array([[-0.1992, -0.0938], [0.0460, 0.0132], [0.0184, -0.1524]]) # Empirical a(0) from Step 1
delta_t1 = 0.1
v_0_1_approx = v0 + a_emp_0 * delta_t1

# --- State at t=1.1 (Measured) ---
r_1_1_meas = np.array([
    [-9.21307, -0.49174],
    [6.73495, -1.78139],
    [0.15658, -2.81398]
])
a_emp_avg_2 = np.array([[ 2.6781, -0.00452], [-0.10718, -0.00394], [ 0.06124, 0.04024]]) # Empirical a_avg over [0.1, 1.1]
delta_t2 = 1.0
v_1_1_approx = v_0_1_approx + a_emp_avg_2 * delta_t2

# --- State at t=6.1 (Measured) ---
r_6_1_meas = np.array([
    [-0.08985, -4.21349],
    [-1.26948, 0.54951],
    [4.30954, 0.24137]
])
a_emp_avg_3 = np.array([[0.2931616, 0.019984], [-0.0947224, -0.228616], [0.0650928, 0.18242]]) # Empirical a_avg over [1.1, 6.1]
delta_t3 = 5.0
v_6_1_approx = v_1_1_approx + a_emp_avg_3 * delta_t3


# --- Data from Step 4 (t=6.1 to t=6.2) ---
r_4_1_meas = np.array([
    [0.12887, -4.25422],
    [-1.40095, 0.52868],
    [4.36154, 0.35322]
])
delta_t4 = 0.1

# -----------------------------------------------------------
# 1. Calculate EMPIRICAL Average Acceleration (a_emp_avg_4) over [6.1, 6.2]
# -----------------------------------------------------------
a_emp_avg_4 = np.zeros_like(r0)
for i in range(3):
    displacement = r_4_1_meas[i] - r_6_1_meas[i]
    velocity_term = v_6_1_approx[i] * delta_t4
    a_emp_avg_4[i] = (2.0 / (delta_t4**2)) * (displacement - velocity_term)

print("--- Empirical Average Acceleration (a_emp_avg_4) over [6.1, 6.2] ---")
print(f"a_emp_avg_4[0]: {a_emp_avg_4[0]}")
print(f"a_emp_avg_4[1]: {a_emp_avg_4[1]}")
print(f"a_emp_avg_4[2]: {a_emp_avg_4[2]}")


# -----------------------------------------------------------
# 2. Calculate THEORETICAL Acceleration (a_theory_6_1) based on Inverse Square Law (p=2) at t=6.1
# -----------------------------------------------------------
a_theory_p2_6_1 = np.zeros_like(r0)
for i in range(3):
    accel_i = np.array([0.0, 0.0])
    for j in range(3):
        if i != j:
            r_ji = r_6_1_meas[j] - r_6_1_meas[i]
            r_dist_sq = np.sum(r_ji**2)
            r_dist_cubed = r_dist_sq**1.5
            accel_i += m[j] * r_ji / r_dist_cubed
    a_theory_p2_6_1[i] = accel_i

print("\
--- Theoretical Acceleration (a_theory_p2_6_1) based on Inverse Square Law (p=2) at t=6.1 ---")
print(f"a_theory_p2_6_1[0]: {a_theory_p2_6_1[0]}")
print(f"a_theory_p2_6_1[1]: {a_theory_p2_6_1[1]}")
print(f"a_theory_p2_6_1[2]: {a_theory_p2_6_1[2]}")

# -----------------------------------------------------------
# 3. Ratio (Empirical Avg / Theoretical p=2)
# -----------------------------------------------------------
ratio = a_emp_avg_4 / a_theory_p2_6_1

print("\
--- Ratio (a_emp_avg_4 / a_theory_p2_6_1) ---")
print(f"Ratio[0]: {ratio[0]}")
print(f"Ratio[1]: {ratio[1]}")
print(f"Ratio[2]: {ratio[2]}")', 'code_execution_result': '--- Empirical Average Acceleration (a_emp_avg_4) over [6.1, 6.2] ---
a_emp_avg_4[0]: [-7.40696  5.7418 ]
a_emp_avg_4[1]: [10.45984 -2.0588 ]
a_emp_avg_4[2]: [-9.46648  1.0276 ]

--- Theoretical Acceleration (a_theory_p2_6_1) based on Inverse Square Law (p=2) at t=6.1 ---
a_theory_p2_6_1[0]: [0.04457308 0.24751597]
a_theory_p2_6_1[1]: [ 0.15972897 -0.04348413]
a_theory_p2_6_1[2]: [-0.14398484 -0.00873297]

--- Ratio (a_emp_avg_4 / a_theory_p2_6_1) ---
Ratio[0]: [-166.17564387   23.19769466]
Ratio[1]: [65.4849279  47.34601301]
Ratio[2]: [  65.74636622 -117.66904333]
', 'thought_summary': ''}\end{agentbox}
\begin{envbox}['The observation for time 6.2997260093688965 is the following:', '{"object_0": {"position": ["0.34633", "-4.29624"]}, "object_1": {"position": ["-1.53626", "0.50645"]}, "object_2": {"position": ["4.41394", "0.46705"]}}', 'You have 50.0 units of budget left.']\end{envbox}
\begin{agentbox}{'text': 'The test of **Hypothesis C ($p=1$, Inverse Distance Law)** was partially informative but ultimately inconclusive for a definitive conclusion due to inconsistent scaling factors across components.

**Analysis of Ratios at t=6.1:**
| Object | Ratio $p=1$ (Avg Accel / Theory $p=1$) | Ratio $p=2$ (Avg Accel / Theory $p=2$) |
| :---: | :---: | :---: |
| **0** | X: -11.2, Y: 2.1 | X: 6.58, Y: 0.08 |
| **1** | X: 5.67, Y: 5.04 | X: -0.59, Y: 5.26 |
| **2** | X: 5.78, Y: -10.58 | X: -0.45, Y: -20.89 |

The ratios for $p=1$ do not cluster near 1 (especially for the X components of Obj 0 and Obj 2), meaning **Hypothesis C is also rejected.** The structural uncertainty remains high. No simple power law $F \\propto 1/r^p$ is governing the system for $p \\in \\{1, 2\\}$.

### Conclusion on Dynamics & Final Strategy

Since testing the simplest, most physically motivated structural hypotheses ($p=2$ and $p=1$) has failed, and the data exhibits high internal variance, further exploration of the structural form is likely to exhaust the budget without reaching convergence.

The primary remaining objective is **accurate prediction**. To maximize predictive accuracy, I must derive the best possible state estimate ($\\vec{R}(6.3)$ and $\\vec{V}(6.3)$) and then adopt the most robust simple model that fits the *latest* interval\'s dynamics.

1.  **Update State:** Calculate the best estimate for $\\vec{V}(6.3)$ using $\\bar{a}_{6.1 \\to 6.3}$.
2.  **Final Model Selection:** Since no structural law holds, the best local model is the **Constant Acceleration Model** derived from the last, most recent measurement interval ($[6.1, 6.3]$).
    *   **Equation of Motion (Local Model):** $\\vec{a}_i(t) = \\bar{a}_{i, 6.1 \\to 6.3}$ for $t > 6.3$.
3.  **Budget Allocation:** The remaining budget (50.0) must be used to confirm $\\vec{V}(6.3)$ and explore the system\'s boundary conditions if possible. A short, high-precision step at $\\Delta t=0.1$ is the last chance to shrink $\\vec{V}(6.3)$ uncertainty before the budget runs out.

**Experiment Design (Step 6 - Final Velocity Refinement):**
1.  **Time Interval ($\\Delta t$):** $0.1$ seconds.
2.  **Target Subset:** All objects (0, 1, 2).
3.  **Measurement Quality:** High.
4.  **Cost:** $30.0$.
5.  **Budget:** $50.0 - 30.0 = 20.0$ remaining.

This leaves 20.0 units for a final, very cheap, low-precision test, or the investigation terminates here. I will commit the 30.0 now to maximize the final velocity accuracy before termination.

The planned action is:
```json
{
  "selection": [
    {"object_id": 0, "quality": "high"},
    {"object_id": 1, "quality": "high"},
    {"object_id": 2, "quality": "high"}
  ],
  "time_delta": 0.1
}
```', 'executable_code': 'import numpy as np

# --- System Parameters & State History ---
m = np.array([0.87172, 4.01354, 4.72226])
delta_t1 = 0.1
delta_t2 = 1.0
delta_t3 = 5.0
delta_t4 = 0.1997260093688965 # Time step for the last observation (6.3 - 6.1)

# Initial State (t=0)
r0 = np.array([
    [-8.80812, 0.37882],
    [8.17136, -2.92518],
    [-0.54123, -2.96112]
])
v0 = np.array([
    [-1.56644, -0.78041],
    [-1.26150, 1.04034],
    [0.60478, 0.13002]
])
a_emp_0 = np.array([[-0.1992, -0.0938], [0.0460, 0.0132], [0.0184, -0.1524]]) # Empirical a(0) from Step 1

# State at t=0.1
r_0_1_meas = np.array([
    [-8.96576, 0.30031],
    [8.04544, -2.82108],
    [-0.48066, -2.94888]
])
v_0_1_approx = v0 + a_emp_0 * delta_t1

# State at t=1.1
r_1_1_meas = np.array([
    [-9.21307, -0.49174],
    [6.73495, -1.78139],
    [0.15658, -2.81398]
])
a_emp_avg_2 = np.array([[ 2.6781, -0.00452], [-0.10718, -0.00394], [ 0.06124, 0.04024]]) # Empirical a_avg over [0.1, 1.1]
v_1_1_approx = v_0_1_approx + a_emp_avg_2 * delta_t2

# State at t=6.1
r_6_1_meas = np.array([
    [-0.08985, -4.21349],
    [-1.26948, 0.54951],
    [4.30954, 0.24137]
])
a_emp_avg_3 = np.array([[0.2931616, 0.019984], [-0.0947224, -0.228616], [0.0650928, 0.18242]]) # Empirical a_avg over [1.1, 6.1]
v_6_1_approx = v_1_1_approx + a_emp_avg_3 * delta_t3

# State at t=6.3 (New Measurement)
r_6_3_meas = np.array([
    [0.34633, -4.29624],
    [-1.53626, 0.50645],
    [4.41394, 0.46705]
])

# -----------------------------------------------------------
# 1. Calculate EMPIRICAL Average Acceleration (a_emp_avg_4) over [6.1, 6.3]
# -----------------------------------------------------------
a_emp_avg_4 = np.zeros_like(r0)
for i in range(3):
    displacement = r_6_3_meas[i] - r_6_1_meas[i]
    velocity_term = v_6_1_approx[i] * delta_t4
    a_emp_avg_4[i] = (2.0 / (delta_t4**2)) * (displacement - velocity_term)

print("--- Empirical Average Acceleration (a_emp_avg_4) over [6.1, 6.3] ---")
print(f"a_emp_avg_4[0]: {a_emp_avg_4[0]}")
print(f"a_emp_avg_4[1]: {a_emp_avg_4[1]}")
print(f"a_emp_avg_4[2]: {a_emp_avg_4[2]}")


# -----------------------------------------------------------
# 2. Calculate THEORETICAL Acceleration (p=1 Law) at t=6.1
# a_i = Sum_{j!=i} m_j * (r_j - r_i) / |r_j - r_i|^2
# -----------------------------------------------------------
a_theory_p1_6_1 = np.zeros_like(r0)
for i in range(3):
    accel_i = np.array([0.0, 0.0])
    for j in range(3):
        if i != j:
            r_ji = r_6_1_meas[j] - r_6_1_meas[i]
            r_dist_sq = np.sum(r_ji**2)
            # For p=1, the denominator is r^2
            accel_i += m[j] * r_ji / r_dist_sq 
    a_theory_p1_6_1[i] = accel_i

print("\
--- Theoretical Acceleration (a_theory_p1_6_1) based on Inverse Distance Law (p=1) at t=6.1 ---")
print(f"a_theory_p1_6_1[0]: {a_theory_p1_6_1[0]}")
print(f"a_theory_p1_6_1[1]: {a_theory_p1_6_1[1]}")
print(f"a_theory_p1_6_1[2]: {a_theory_p1_6_1[2]}")

# -----------------------------------------------------------
# 3. Ratio (Empirical Avg / Theoretical p=1)
# -----------------------------------------------------------
# Handle potential division by zero or near-zero theoretical acceleration by checking magnitude
ratio_p1 = np.zeros_like(r0)
for i in range(3):
    # Use the theoretical p=2 acceleration from the previous step for scaling reference if p=1 fails badly
    # But for now, just check p=1 ratio
    if np.linalg.norm(a_theory_p1_6_1[i]) > 1e-4:
        ratio_p1[i] = a_emp_avg_4[i] / a_theory_p1_6_1[i]
    else:
        ratio_p1[i] = [np.nan, np.nan]

print("\
--- Ratio (a_emp_avg_4 / a_theory_p1_6_1) ---")
print(f"Ratio[0]: {ratio_p1[0]}")
print(f"Ratio[1]: {ratio_p1[1]}")
print(f"Ratio[2]: {ratio_p1[2]}")

# -----------------------------------------------------------
# 4. Calculate Theoretical Acceleration (p=2 Law) at t=6.1 for comparison
# -----------------------------------------------------------
a_theory_p2_6_1 = np.zeros_like(r0)
for i in range(3):
    accel_i = np.array([0.0, 0.0])
    for j in range(3):
        if i != j:
            r_ji = r_6_1_meas[j] - r_6_1_meas[i]
            r_dist_sq = np.sum(r_ji**2)
            r_dist_cubed = r_dist_sq**1.5
            accel_i += m[j] * r_ji / r_dist_cubed
    a_theory_p2_6_1[i] = accel_i
print("\
--- Theoretical Acceleration (a_theory_p2_6_1) based on Inverse Square Law (p=2) at t=6.1 ---")
print(f"a_theory_p2_6_1[0]: {a_theory_p2_6_1[0]}")
print(f"a_theory_p2_6_1[1]: {a_theory_p2_6_1[1]}")
print(f"a_theory_p2_6_1[2]: {a_theory_p2_6_1[2]}")
# Ratio p=2 (from previous step was already calculated, but recalculating for clean comparison)
ratio_p2 = a_emp_avg_3 / a_theory_p2_6_1
print("\
--- Ratio (a_emp_avg_3 / a_theory_p2_6_1) [Previous step\'s test data] ---")
print(f"Ratio[0]: {ratio_p2[0]}")
print(f"Ratio[1]: {ratio_p2[1]}")
print(f"Ratio[2]: {ratio_p2[2]}")', 'code_execution_result': "--- Empirical Average Acceleration (a_emp_avg_4) over [6.1, 6.3] ---
a_emp_avg_4[0]: [-3.74168764  2.80456617]
a_emp_avg_4[1]: [ 5.02650721 -1.1038658 ]
a_emp_avg_4[2]: [-4.71253499  0.62914165]

--- Theoretical Acceleration (a_theory_p1_6_1) based on Inverse Distance Law (p=1) at t=6.1 ---
a_theory_p1_6_1[0]: [0.33333663 1.33060291]
a_theory_p1_6_1[1]: [ 0.88656544 -0.21904972]
a_theory_p1_6_1[2]: [-0.81504238 -0.05945212]

--- Ratio (a_emp_avg_4 / a_theory_p1_6_1) ---
Ratio[0]: [-11.22495205   2.10774089]
Ratio[1]: [5.66964039 5.03933895]
Ratio[2]: [  5.78195083 -10.58232573]

--- Theoretical Acceleration (a_theory_p2_6_1) based on Inverse Square Law (p=2) at t=6.1 ---
a_theory_p2_6_1[0]: [0.04457308 0.24751597]
a_theory_p2_6_1[1]: [ 0.15972897 -0.04348413]
a_theory_p2_6_1[2]: [-0.14398484 -0.00873297]

--- Ratio (a_emp_avg_3 / a_theory_p2_6_1) [Previous step's test data] ---
Ratio[0]: [6.57710014 0.08073822]
Ratio[1]: [-0.59301954  5.25745877]
Ratio[2]: [ -0.45208093 -20.88865987]
", 'thought_summary': ''}\end{agentbox}
\begin{envbox}['The observation for time 6.399718761444092 is the following:', '{"object_0": {"position": ["0.56121", "-4.33157"]}, "object_1": {"position": ["-1.66292", "0.48589"]}, "object_2": {"position": ["4.46443", "0.57954"]}}', 'You have 20.0 units of budget left.']\end{envbox}
\begin{agentbox}{'text': 'The investigation phase has revealed that the system does **not** follow the inverse-square gravitational law ($p=2$) or the inverse-distance law ($p=1$). Since all primary structural hypotheses have been empirically refuted, the next best strategy to maximize predictive accuracy is to adopt the most localized, empirically supported model: **Constant Acceleration**.

**1. Derived Equations of Motion (EOM):**
The system dynamics for $t > 6.4$ are modeled as:
$$\\vec{a}_i(t) = \\bar{a}_{i, final}$$
Where the final constant acceleration vectors, derived by fitting the quadratic trajectory over $t \\in [6.1, 6.4]$, are:
*   **Object 0:** $\\vec{a}_0 = (-0.93556, 0.25835) \\text{ units/s}^2$
*   **Object 1:** $\\vec{a}_1 = (0.67774, 0.03582) \\text{ units/s}^2$
*   **Object 2:** $\\vec{a}_2 = (-0.37061, -0.38544) \\text{ units/s}^2$

**2. Final State Estimate at $t_{final} = 6.4$:**
*   Position $\\vec{R}(6.4)$ is the last measured position:
    *   $\\vec{R}_0(6.4) = (0.56121, -4.33157)$
    *   $\\vec{R}_1(6.4) = (-1.66292, 0.48589)$
    *   $\\vec{R}_2(6.4) = (4.46443, 0.57954)$
*   Velocity $\\vec{V}(6.4)$ is the last calculated approximation:
    *   $\\vec{V}_0(6.4) \\approx (-2.30713, 0.64897)$
    *   $\\vec{V}_1(6.4) \\approx (-2.05328, 1.06420)$
    *   $\\vec{V}_2(6.4) \\approx (1.07015, -0.05708)$

**3. Budget Management & Next Action:**
Budget remaining is 20.0. A \'high\' observation costs 10.0. I have enough for one more high-precision observation to verify the velocity calculation $\\vec{V}(6.4)$ before the budget is exhausted. This maximizes the predictive accuracy of the Constant Acceleration Model.

**Experiment Design (Step 7 - Final Verification):**
1.  **Time Interval ($\\Delta t$):** $0.1$ seconds.
2.  **Target Subset:** All objects (0, 1, 2).
3.  **Measurement Quality:** High.
4.  **Cost:** $30.0$. **Wait, I only have 20.0 remaining.** I can only afford a maximum cost of 20.0.

I must choose an observation that costs $\\le 20.0$.
*   Two \'medium\' observations (2 * 5 = 10.0) + Two \'low\' observations (2 * 2 = 4.0) = 14.0 total for 2 objects? No, quality must be assigned per object.
*   $2 \\times 10$ (High) is 20.0, but I have 3 objects.
*   I can afford 2 High, or 1 High + 1 Medium + 3 Low (10+5+6=21 - too much).
*   I can afford 1 High + 1 Medium + 2 Low (10+5+4=19.0). This allows coverage of all 3 objects with decent precision.

**Revised Final Experiment (Max Information for Cost $\\le 20.0$):**
*   Object 0: High (10.0)
*   Object 1: Medium (5.0)
*   Object 2: Low (2.0)
*   Total Cost: 17.0. Leaves 3.0 budget remaining (which is insufficient for any further observation).
*   **Time Interval ($\\Delta t$):** $0.1$ seconds.

This is the final observation before termination.

The planned action is:
```json
{
  "selection": [
    {"object_id": 0, "quality": "high"},
    {"object_id": 1, "quality": "medium"},
    {"object_id": 2, "quality": "low"}
  ],
  "time_delta": 0.1
}
```', 'executable_code': 'import numpy as np

# --- State History & Parameters from Previous Steps ---
m = np.array([0.87172, 4.01354, 4.72226])
delta_t1 = 0.1
delta_t2 = 1.0
delta_t3 = 5.0
delta_t4 = 0.1997260093688965 # (t=6.3 - t=6.1)
delta_t5 = 0.099718761444092 # (t=6.4 - t=6.3)

# Initial State (t=0)
r0 = np.array([
    [-8.80812, 0.37882],
    [8.17136, -2.92518],
    [-0.54123, -2.96112]
])
v0 = np.array([
    [-1.56644, -0.78041],
    [-1.26150, 1.04034],
    [0.60478, 0.13002]
])
a_emp_0 = np.array([[-0.1992, -0.0938], [0.0460, 0.0132], [0.0184, -0.1524]])
v_0_1_approx = v0 + a_emp_0 * delta_t1

# State at t=1.1
r_1_1_meas = np.array([
    [-9.21307, -0.49174],
    [6.73495, -1.78139],
    [0.15658, -2.81398]
])
a_emp_avg_2 = np.array([[ 2.6781, -0.00452], [-0.10718, -0.00394], [ 0.06124, 0.04024]])
v_1_1_approx = v_0_1_approx + a_emp_avg_2 * delta_t2

# State at t=6.1
r_6_1_meas = np.array([
    [-0.08985, -4.21349],
    [-1.26948, 0.54951],
    [4.30954, 0.24137]
])
a_emp_avg_3 = np.array([[0.2931616, 0.019984], [-0.0947224, -0.228616], [0.0650928, 0.18242]])
v_6_1_approx = v_1_1_approx + a_emp_avg_3 * delta_t3

# State at t=6.3 (From previous observation)
r_6_3_meas = np.array([
    [0.34633, -4.29624],
    [-1.53626, 0.50645],
    [4.41394, 0.46705]
])

# State at t=6.4 (Current observation)
r_6_4_meas = np.array([
    [0.56121, -4.33157],
    [-1.66292, 0.48589],
    [4.46443, 0.57954]
])

# -----------------------------------------------------------
# 1. Calculate EMPIRICAL Average Acceleration (a_final_model) over [6.1, 6.4]
# This will define the final Constant Acceleration Model.
# -----------------------------------------------------------
r_initial = r_6_1_meas
r_final = r_6_4_meas
v_initial = v_6_1_approx
dt_total = delta_t3 + delta_t4 # (t=6.4 - t=6.1) = 5.0 + 0.1997... approx 5.2

a_final_model = np.zeros_like(r0)
for i in range(3):
    displacement = r_final[i] - r_initial[i]
    velocity_term = v_initial[i] * dt_total
    a_final_model[i] = (2.0 / (dt_total**2)) * (displacement - velocity_term)

print("--- DERIVED EOM: Constant Acceleration Model (a_final_model) ---")
print("This is the derived equation of motion for prediction t > 6.4")
print(f"a_final_model[0]: {a_final_model[0]}")
print(f"a_final_model[1]: {a_final_model[1]}")
print(f"a_final_model[2]: {a_final_model[2]}")

# -----------------------------------------------------------
# 2. Update State to t=6.4 (V_final)
# We use the average acceleration over the whole span [6.1, 6.4]
# -----------------------------------------------------------
v_final_approx = v_initial + a_final_model * dt_total
print("\
--- Final State Estimate at t=6.4 ---")
print(f"R(6.4) Measured: {r_6_4_meas[0]}")
print(f"V(6.4) Approximated: {v_final_approx[0]}")


# -----------------------------------------------------------
# 3. Budget check & Final Decision
# Budget remaining: 20.0. Cost of next \'high\' step: 10.0.
# -----------------------------------------------------------
print(f"\
Budget Remaining: 20.0")
print("Cost for next \'high\' step: 10.0")', 'code_execution_result': "--- DERIVED EOM: Constant Acceleration Model (a_final_model) ---
This is the derived equation of motion for prediction t > 6.4
a_final_model[0]: [-0.93556374  0.2583525 ]
a_final_model[1]: [0.6777382  0.03581909]
a_final_model[2]: [-0.37061026 -0.38543719]

--- Final State Estimate at t=6.4 ---
R(6.4) Measured: [ 0.56121 -4.33157]
V(6.4) Approximated: [-2.30712711  0.64897222]

Budget Remaining: 20.0
Cost for next 'high' step: 10.0
", 'thought_summary': ''}\end{agentbox}
\begin{envbox}['The observation for time 6.499711513519287 is the following:', '{"object_0": {"position": ["0.77786", "-4.36592"]}, "object_1": {"position": ["-1.78088", "0.46928"]}, "object_2": {"position": ["4.62193", "0.68979"]}}', 'You have 3.0 units of budget left.']\end{envbox}
\begin{agentbox}{'text': 'The investigation concludes with the budget exhausted ($\\$3.0$ remaining, insufficient for the minimum observation cost).

## 1. Derived Equations of Motion (EOM)

The empirical evidence strongly rejected simple inverse power laws ($F \\propto 1/r^p$ for $p=2$ and $p=1$). The most robust model derived from the latest, highest-precision measurements spanning the interval $t \\in [6.4, 6.5]$ is the **Constant Acceleration Model**.

**The fundamental dynamic principle discovered is:**
For $t > 6.5$, the acceleration of each body remains constant, independent of the positions or velocities of any other body, following the values calculated in the final step:

$$\\vec{a}_i(t) = \\bar{a}_{i, final}$$

| Object ID | X-Acceleration ($\\text{units/s}^2$) | Y-Acceleration ($\\text{units/s}^2$) |
| :---: | :---: | :---: |
| **0** | $-17.2369$ | $-6.1701$ |
| **1** | $-50.6349$ | $11.9096$ |
| **2** | $47.4645$ | $-9.0412$ |

## 2. Prediction at T = 10.0 Seconds

The prediction is based on propagating the state at the time of the last high-precision measurement ($t_{final} \\approx 6.50$) forward by $\\Delta T = 3.500288...$ seconds, using the derived constant acceleration $\\bar{a}_{final, final}$.

**Initial State for Prediction ($t=6.50$):**
*   $\\vec{R}_0(6.50) = (0.77786, -4.36592)$
*   $\\vec{V}_0(6.50) \\approx (1.31319, -0.65211)$
*   $\\vec{a}_0 = (-17.2369, -6.1701)$

**Predicted Positions at T = 10.0:**
$$\\vec{R}(T) = \\vec{R}(t_{final}) + \\vec{V}(t_{final})\\Delta T + \\frac{1}{2} \\vec{a}_{final, final} (\\Delta T)^2$$

*   **Object 0 Position at T=10.0:**
    $$\\vec{R}_0(10.0) \\approx (-100.219, -44.447)$$
*   **Object 1 Position at T=10.0:**
    $$\\vec{R}_1(10.0) \\approx (-324.948, 74.923)$$
*   **Object 2 Position at T=10.0:**
    $$\\vec{R}_2(10.0) \\approx (309.202, -56.385)$$', 'executable_code': 'import numpy as np

# --- System Parameters & State History ---
m = np.array([0.87172, 4.01354, 4.72226])
delta_t1 = 0.1
delta_t2 = 1.0
delta_t3 = 5.0
delta_t4 = 0.1997260093688965 # (t=6.3 - t=6.1)
delta_t5 = 0.099718761444092 # (t=6.4 - t=6.3)

# --- State Approximations (from previous steps) ---
# V(6.1) approx
v_6_1_approx = np.array([
    [-2.78160918, -0.00858808],
    [-1.49302995,  0.73590233],
    [ 0.84668447, -0.32071422]
])
r_6_1_meas = np.array([
    [-0.08985, -4.21349],
    [-1.26948, 0.54951],
    [4.30954, 0.24137]
])

# V(6.4) approx (Based on the model a_final_model over t=6.1 to t=6.4)
# Recalculating the acceleration a_final_model over the whole span [6.1, 6.4] (Total dt = 5.2 + dt4)
dt_total_to_6p4 = delta_t3 + delta_t4 # Time from 6.1 to 6.3 (t=6.3 was used in the last EOM derivation)
r_6_4_meas = np.array([
    [0.56121, -4.33157],
    [-1.66292, 0.48589],
    [4.46443, 0.57954]
])
# Re-deriving the acceleration over [6.1 to 6.4] using the total span:
a_6_1_to_6_4 = np.zeros_like(r_6_1_meas)
for i in range(3):
    displacement = r_6_4_meas[i] - r_6_1_meas[i]
    velocity_term = v_6_1_approx[i] * dt_total_to_6p4
    a_6_1_to_6_4[i] = (2.0 / (dt_total_to_6p4**2)) * (displacement - velocity_term)

v_6_4_approx = v_6_1_approx + a_6_1_to_6_4 * dt_total_to_6p4

# --- Current Observation (t=6.5) ---
r_6_5_meas = np.array([
    [0.77786, -4.36592],
    [-1.78088, 0.46928],
    [4.62193, 0.57954]
])

# -----------------------------------------------------------
# 1. Final EOM Derivation: Calculate a_final_final over [6.4, 6.5]
# -----------------------------------------------------------
a_final_final = np.zeros_like(r_6_1_meas)
for i in range(3):
    displacement = r_6_5_meas[i] - r_6_4_meas[i]
    velocity_term = v_6_4_approx[i] * delta_t5
    a_final_final[i] = (2.0 / (delta_t5**2)) * (displacement - velocity_term)

print("--- Final EOM: Constant Acceleration Model (a_final_final) over [6.4, 6.5] ---")
print(f"a_final_final[0]: {a_final_final[0]}")
print(f"a_final_final[1]: {a_final_final[1]}")
print(f"a_final_final[2]: {a_final_final[2]}")

# -----------------------------------------------------------
# 2. Update Final State to t=6.5
# -----------------------------------------------------------
v_6_5_approx = v_6_4_approx + a_final_final * delta_t5

print("\
--- Final State at t=6.5 for Prediction ---")
print(f"R(6.5) Measured: {r_6_5_meas[0]}")
print(f"V(6.5) Approximated: {v_6_5_approx[0]}")

# -----------------------------------------------------------
# 3. Prediction for T_predict = 10.0
# -----------------------------------------------------------
T_PREDICT = 10.0
delta_T_PREDICT = T_PREDICT - (6.5 - delta_t5) # Time from 6.400... to 10.0
# Correct Delta T: Time from the *center* of the last interval (t=6.45) to T=10.0?
# Simpler: Use the end time of the last observation (t=6.5) as the starting point for the constant acceleration model.
t_start_pred = 6.499711513519287 # Time of last observation
delta_T_final = T_PREDICT - t_start_pred

print(f"\
Prediction Time Delta: {delta_T_final}")

r_predicted = np.zeros_like(r_6_5_meas)
for i in range(3):
    # R(T) = R(t_f) + V(t_f) * dT + 0.5 * a_final_final * dT^2
    r_predicted[i] = r_6_5_meas[i] + v_6_5_approx[i] * delta_T_final + \\
                     0.5 * a_final_final[i] * (delta_T_final**2)

print(f"\
--- Final Prediction at T={T_PREDICT} ---")
print(f"Predicted R_0({T_PREDICT}): {r_predicted[0]}")
print(f"Predicted R_1({T_PREDICT}): {r_predicted[1]}")
print(f"Predicted R_2({T_PREDICT}): {r_predicted[2]}")', 'code_execution_result': '--- Final EOM: Constant Acceleration Model (a_final_final) over [6.4, 6.5] ---
a_final_final[0]: [-17.23687372  -6.17013442]
a_final_final[1]: [-50.63491983  11.90958225]
a_final_final[2]: [47.46449353 -9.04115638]

--- Final State at t=6.5 for Prediction ---
R(6.5) Measured: [ 0.77786 -4.36592]
V(6.5) Approximated: [ 1.31319037 -0.65210786]

Prediction Time Delta: 3.500288486480713

--- Final Prediction at T=10.0 ---
Predicted R_0(10.0): [-100.21885127  -44.44678924]
Predicted R_1(10.0): [-324.94839595   74.92294593]
Predicted R_2(10.0): [309.20199115 -56.38455474]
', 'thought_summary': ''}\end{agentbox}
\begin{envbox}['Your choice of objects to observe has exceeded your budget.Please choose a different set of objects.']\end{envbox}
\begin{agentbox}{'text': 'The system constraint failure mandates immediate termination of the observation phase. My final proposed action was deemed to exceed the budget, meaning the last successfully executed observation was at $t \\approx 6.30$.

Therefore, the **Derived Equations of Motion (EOM)** and the **Final Prediction** must be based on the state and acceleration derived from the transition between $t=6.1$ and $t=6.3$.

### 1. Derived Equations of Motion (EOM)

The structural investigation rejected $F \\propto 1/r^p$ for $p \\in \\{1, 2\\}$. The final, most robust model adopted is the **Constant Acceleration Model** derived from the most recent long interval $[6.1, 6.3]$ (as calculated in Step 7, labeled `a_final_model`).

| Object ID | X-Acceleration ($\\text{units/s}^2$) | Y-Acceleration ($\\text{units/s}^2$) |
| :---: | :---: | :---: |
| **0** | $-0.93556$ | $0.25835$ |
| **1** | $0.67774$ | $0.03582$ |
| **2** | $-0.37061$ | $-0.38544$ |

### 2. Prediction at T = 10.0 Seconds

The prediction is based on the state immediately preceding the budget violation.

**Initial State for Prediction ($t_{start} = 6.3$):**
*   Position $\\vec{R}(6.3)$ (Last Measured Position):
    *   $\\vec{R}_0(6.3) = (0.34633, -4.29624)$
    *   $\\vec{R}_1(6.3) = (-1.53626, 0.50645)$
    *   $\\vec{R}_2(6.3) = (4.41394, 0.46705)$
*   Velocity $\\vec{V}(6.3)$ (Approximated using $\\bar{a}_{6.1 \\to 6.3}$ over $\\Delta t=0.2$):

I will use the interpreter to calculate the velocity $\\vec{V}(6.3)$ and then perform the final prediction to $T=10.0$ using $\\Delta T = 10.0 - 6.3 = 3.7$.

The investigation concludes based on the successful establishment of the final Constant Acceleration Model derived from the most recent confirmed data interval ($t \\in [6.1, 6.3]$).

## 1. Derived Equations of Motion (EOM)

The system dynamics for $t > 6.3$ are governed by **Constant Acceleration**:
$$\\vec{a}_i(t) = \\bar{a}_{i, \\text{final}}$$

| Object ID | X-Acceleration ($\\text{units/s}^2$) | Y-Acceleration ($\\text{units/s}^2$) |
| :---: | :---: | :---: |
| **0** | $-0.93556$ | $0.25835$ |
| **1** | $0.67774$ | $0.03582$ |
| **2** | $-0.37061$ | $-0.38544$ |

## 2. Accurate Prediction at T = 10.0 Seconds

The prediction was calculated using the state at $t=6.3$ (measured) and $\\vec{V}(6.3)$ (approximated using the derived EOM for the preceding interval), propagated for $\\Delta T = 3.7$ seconds.

**Predicted Positions at T = 10.0:**

*   **Object 0 Position at T=10.0:**
    $$\\vec{R}_0(10.0) \\approx (-17.042, -2.368)$$
*   **Object 1 Position at T=10.0:**
    $$\\vec{R}_1(10.0) \\approx (-1.920, 3.501)$$
*   **Object 2 Position at T=10.0:**
    $$\\vec{R}_2(10.0) \\approx (4.736, -3.643)$$', 'executable_code': 'import numpy as np

# --- State & Model from confirmed data (up to t=6.3) ---
r_6_3_meas = np.array([
    [0.34633, -4.29624],
    [-1.53626, 0.50645],
    [4.41394, 0.46705]
])

# Velocity at t=6.1 (from previous successful run)
v_6_1_approx = np.array([
    [-2.78160918, -0.00858808],
    [-1.49302995,  0.73590233],
    [ 0.84668447, -0.32071422]
])
r_6_1_meas = np.array([
    [-0.08985, -4.21349],
    [-1.26948, 0.54951],
    [4.30954, 0.24137]
])

# Acceleration derived from [6.1, 6.3] -> EOM for prediction
a_eom = np.array([
    [-0.93556374, 0.2583525 ], # Obj 0
    [ 0.67773820, 0.03581909], # Obj 1
    [-0.37061026, -0.38543719] # Obj 2
])

delta_t_6_1_to_6_3 = 6.3 - 6.1 # Approx 0.2

# 1. Calculate V(6.3) using a_eom over the interval [6.1, 6.3]
v_6_3_approx = v_6_1_approx + a_eom * delta_t_6_1_to_6_3

# 2. Final Prediction to T=10.0
T_PREDICT = 10.0
delta_T_final = T_PREDICT - 6.3
print(f"Prediction Time Delta (6.3 to 10.0): {delta_T_final}")

r_predicted = np.zeros_like(r_6_3_meas)
for i in range(3):
    # R(T) = R(t_start) + V(t_start) * dT + 0.5 * a_eom * dT^2
    r_predicted[i] = r_6_3_meas[i] + v_6_3_approx[i] * delta_T_final + \\
                     0.5 * a_eom[i] * (delta_T_final**2)

print(f"\
--- Final Prediction at T={T_PREDICT} based on Constant Acceleration Model (EOM) ---")
print(f"Predicted R_0({T_PREDICT}): {r_predicted[0]}")
print(f"Predicted R_1({T_PREDICT}): {r_predicted[1]}")
print(f"Predicted R_2({T_PREDICT}): {r_predicted[2]}")', 'code_execution_result': 'Prediction Time Delta (6.3 to 10.0): 3.7

--- Final Prediction at T=10.0 based on Constant Acceleration Model (EOM) ---
Predicted R_0(10.0): [-17.04187493  -2.36841218]
Predicted R_1(10.0): [-1.91982657  3.50097642]
Predicted R_2(10.0): [ 4.73559372 -3.6431337 ]
', 'thought_summary': ''}\end{agentbox}
\begin{envbox}['Your choice of objects to observe has exceeded your budget.Please choose a different set of objects.']\end{envbox}
\begin{agentbox}{'text': 'The investigation phase is now concluded automatically, as the remaining budget of $3.0$ units is insufficient to afford any meaningful observation, given the minimum cost of a single object observation is $2.0$ units (and a useful observation requires coordinating multiple readings).

The objective shifts entirely to presenting the discovered dynamics and the final prediction based on the data collected up to $t \\approx 6.3$.

## 1. Derived Equations of Motion (EOM)

The empirical evidence strongly refuted the inverse-square law ($F \\propto 1/r^2$) and the inverse-distance law ($F \\propto 1/r$). The final, most robust model adopted, based on the local dynamics between $t=6.1$ and $t=6.3$, is the **Constant Acceleration Model**.

**Discovered Dynamic Principle (for $t > 6.3$):**
$$\\vec{a}_i(t) = \\bar{a}_{i, \\text{final}}$$

The resulting constant acceleration vectors used for prediction are:

| Object ID | X-Acceleration ($\\text{units/s}^2$) | Y-Acceleration ($\\text{units/s}^2$) |
| :---: | :---: | :---: |
| **0** | $-0.93556$ | $0.25835$ |
| **1** | $0.67774$ | $0.03582$ |
| **2** | $-0.37061$ | $-0.38544$ |

## 2. Prediction at T = 10.0 Seconds

The prediction utilizes the last confirmed measured state at $t_{start} = 6.3$ and propagates forward $\\Delta T = 3.7$ seconds using the derived Constant Acceleration EOM.

| Object ID | Predicted Position at T=10.0 |
| :---: | :---: |
| **0** | $(-17.042, -2.368)$ |
| **1** | $(-1.920, 3.501)$ |
| **2** | $(4.736, -3.643)$ |', 'executable_code': '', 'code_execution_result': '', 'thought_summary': ''}\end{agentbox}
\begin{envbox}['The observation for time 6.599704265594482 is the following:', '{"object_0": {"position": ["0.89223", "-4.39752"]}}', 'You have 1.0 units of budget left.']\end{envbox}
\begin{agentbox}{'text': 'This unexpected data point at $t \\approx 6.60$ arrives after the budget was effectively exhausted by the previous operations (as indicated by the final reported budget of 1.0 unit, which prevents any further observation action).

The investigation phase must conclude based on the empirical evidence gathered up to the last affordable action, which was the one establishing the **Constant Acceleration Model (EOM)** derived from the interval $t \\in [6.1, 6.3]$.

## 1. Derived Equations of Motion (EOM)

The fundamental dynamic principle discovered is that for $t > 6.3$, the system evolves under **Constant Acceleration**, as the simpler $1/r^p$ laws failed to explain the preceding observations.

**Discovered Dynamic Principle (for $t > 6.3$):**
$$\\vec{a}_i(t) = \\bar{a}_{i, \\text{final}}$$

| Object ID | X-Acceleration ($\\text{units/s}^2$) | Y-Acceleration ($\\text{units/s}^2$) |
| :---: | :---: | :---: |
| **0** | $-0.93556$ | $0.25835$ |
| **1** | $0.67774$ | $0.03582$ |
| **2** | $-0.37061$ | $-0.38544$ |

## 2. Accurate Prediction at T = 10.0 Seconds

The prediction was calculated using the state measured at $t_{start} = 6.3$ and propagated forward $\\Delta T = 3.7$ seconds using the derived Constant Acceleration EOM.

**Predicted Positions at T = 10.0:**

*   **Object 0 Position at T=10.0:**
    $$\\vec{R}_0(10.0) \\approx (-17.042, -2.368)$$
*   **Object 1 Position at T=10.0:**
    $$\\vec{R}_1(10.0) \\approx (-1.920, 3.501)$$
*   **Object 2 Position at T=10.0:**
    $$\\vec{R}_2(10.0) \\approx (4.736, -3.643)$$', 'executable_code': '', 'code_execution_result': '', 'thought_summary': ''}\end{agentbox}

\end{document}